\documentclass[runningheads]{llncs}
\usepackage{graphicx}

\usepackage{tikz}
\usepackage{comment}
\usepackage{amsmath,amssymb} 
\usepackage{color}
\usepackage{soul}

\usepackage[accsupp]{axessibility}  

\usepackage{xcolor-material}
\usepackage{soul}
\usepackage{color}
\usepackage[normalem]{ulem}
\useunder{\uline}{\ul}{}
\usepackage{multirow}
\usepackage{siunitx}    
\usepackage{footnote}
\usepackage[symbol]{footmisc}
\usepackage{capt-of}
\usepackage{xspace}

\newcommand{\dataname} {CHICO\xspace}
\newcommand{\gcnname} {SeS-GCN\xspace}

\DeclareRobustCommand{\myparagraph}[1]
{\noindent\textbf{#1}} 

\usepackage{graphicx}
\usepackage{amsmath}
\usepackage{amssymb}
\usepackage{booktabs}

\usepackage[pagebackref,breaklinks,colorlinks]{hyperref}

\usepackage[capitalize]{cleveref}
\crefname{section}{Sec.}{Secs.}
\Crefname{section}{Section}{Sections}
\Crefname{table}{Table}{Tables}
\crefname{table}{Tab.}{Tabs.}

\title{Anticipating human motion for human-robot collaboration}

\author{First Author\\
Institution1\\
Institution1 address\\
{\tt\small firstauthor@i1.org}
\and
Second Author\\
Institution2\\
First line of institution2 address\\
{\tt\small secondauthor@i2.org}
}

\begin{document}

\pagestyle{headings}
\mainmatter
\def\ECCVSubNumber{5644}  

\title{Pose Forecasting in Industrial Human-Robot Collaboration}

\titlerunning{Pose Forecasting in Industrial Human-Robot Collaboration}
%

\author{
Alessio Sampieri\inst{1} 
\and
Guido Maria D'Amely di Melendugno\inst{1}\index{D'Amely di Melendugno, Guido Maria} 
\and
Andrea Avogaro\inst{2}
\and
Federico Cunico\inst{2}
\and
Francesco Setti\inst{2} 
\and
Geri Skenderi\inst{2}
\and
Marco Cristani\inst{2}
\and
Fabio Galasso\inst{1}
}

\authorrunning{A. Sampieri et al.}
%
\institute{
Sapienza University of Rome\\
\email{\{sampieri, damely, galasso\}@di.uniroma1.it}\\
\and
University of Verona\\
\email{\{andrea.avogaro, federico.cunico, francesco.setti, geri.skenderi, marco.cristani\}@univr.it}
}

\maketitle

\begin{abstract}
Pushing back the frontiers of collaborative robots in industrial environments, we propose a new Separable-Sparse Graph Convolutional Network (\gcnname) for pose forecasting. For the first time, \gcnname{} bottlenecks the interaction of the spatial, temporal and channel-wise dimensions in GCNs, and it learns sparse adjacency matrices by a teacher-student framework.
Compared to the state-of-the-art, it only uses 1.72\% of the parameters and it is $\sim$4 times faster, while still performing comparably in forecasting accuracy on Human3.6M at 1 second in the future, which enables cobots to be aware of human operators.
As a second contribution, we present a new benchmark of Cobots and Humans in Industrial COllaboration (\dataname{}).
\dataname{} includes multi-view videos, 3D poses and trajectories of 20 human operators and cobots, engaging in 7 realistic industrial actions. Additionally, it reports 226 genuine collisions, taking place during the human-cobot interaction.
We test \gcnname{} on \dataname{} for two important perception tasks in robotics: human pose forecasting, where it reaches an average error of 85.3 mm (MPJPE) at 1 sec in the future with a run time of 2.3 msec, and collision detection, by comparing the forecasted human motion with the known cobot motion, obtaining an F1-score of 0.64.\\

\keywords{Human Pose Forecasting, Graph Convolutional Networks, Human-Robot Collaboration in Industry.}

\end{abstract}


\section{Introduction}
\label{sec:introduction}

\begin{figure}[ht]
    \centering
    
    \includegraphics[width=\linewidth]{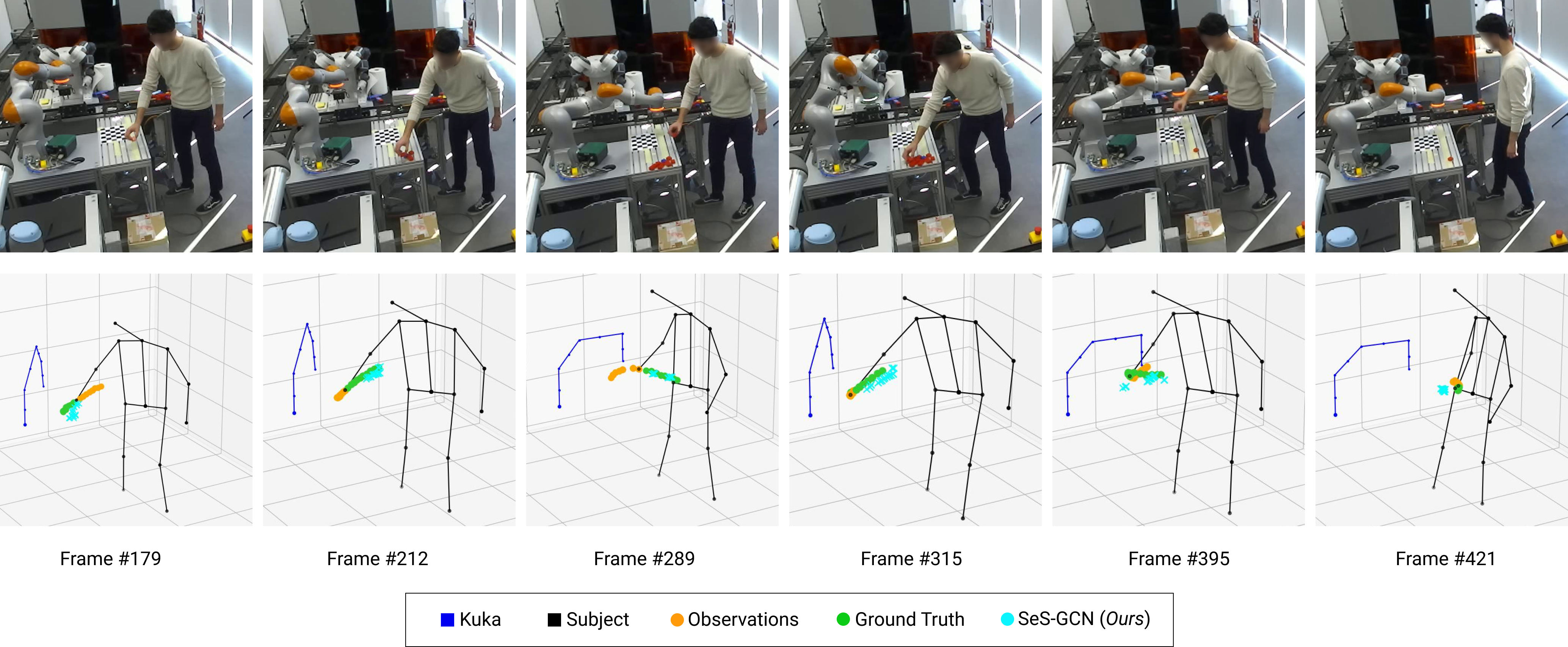}
    \caption{A collision example from our \dataname{} dataset. On the top row some frames of the \emph{Lightweight pick and place} action captured by one of the three cameras. On the bottom row, operator + robot skeletons. The forecasting takes an observation sequence (in yellow, here pictured for the right wrist only), and performs a prediction (cyan) which is compared with the ground truth (green). On frame 395 is it easy to see the robot hitting the operator, which is retracting, as it is evident in frame 421.
    See how the predictions by SeS-GCN follow closely the GT, except during the collision. At collision time, due to the impact, the abrupt change of the arm motion produces uncertain predictions, as it shows from the very irregular predicted trajectory.
    }
    \label{fig:teaser}

\vspace{-0.5cm}
\end{figure}

Collaborative robots (cobots) and modern Human Robot Collaboration (HRC) depart from the traditional separation of functions of industrial robots~\cite{Stein2020}, 
because of the shared workspace~\cite{iso15066}. Additionally cobots and humans perform actions concurrently and they will therefore physically engage in contact. While there is a clear advantage in increased productivity~\cite{Rodriguez-Guerra2021} (improved by as much as 85\%~\cite{Shah2011}) due to the minimization of idle times, there are challenges in the workplace safety~\cite{Gualtieri2020}: it is not about whether there will be contact, but rather about understanding its consequences~\cite{Matthias2016}.

The pioneering work of Shah et al.~\cite{Shah2011} has already shown that, in order to seamlessly and efficiently interact with human co-workers, cobots need to abide by two collaborative principles: (1) Making decisions on-the-fly, and (2) Considering the consequences of their decision on their teammates. The first calls for promptly and accurately detecting the human motion in the workspace. The second principle implies that cobots need to anticipate pose trajectories of their human co-workers and predict future collisions.

Motivated by these problems, the first contribution of our work is a novel Separable-Sparse Graph Convolutional Neural Network (\gcnname{}) for human pose forecasting. Pose forecasting requires understanding of the complex spatio-temporal joint dynamics of the human body and recent trends have highlighted the promises of modelling body kinematics within a single GCN framework \cite{cui20,Dang21,li20,li21,Mao19LTD,Wang_21,Zhao20}.
We have designed \gcnname{} with performance and efficiency in mind, by bringing together, for the first time, three main modelling principles: depthwise-separable graph convolutions~\cite{lai18}, space-time separable graph adjacency matrices ~\cite{stsgcn}, and sparse graph adjacency matrices~\cite{shi21}. 
In \gcnname{}, \emph{separable} stands for limiting the interplay of joints with others (space), at different frames (time) and per channel (depth-wise). Within the GCN, different channels, frames and joints still interact by means of multi-hop messages
For the first time, sparsity is achieved by a teacher-student framework.
The reduced interaction and sparsity results in comparable or less parameters than all GCN-based baselines~\cite{lai18,stsgcn,shi21}, while improving performance by at least 2.7\%. 
Compared to the state-of-the-art (SoA)~\cite{Mao20DCT}, \gcnname{} is lightweight, only using 1.72\% of its parameters, it is $\sim$4 times faster, while remaining competitive with just 1.5\% larger error on Human 3.6M~\cite{h36m_pami} when predicting 1 sec in the future.
The model is described in detail in Sec.~\ref{sec:methodology}, experiments and ablation studies are illustrated in  Sec.~\ref{sec:experimental_results_h36m}.

We also introduce the very first benchmark of Cobots and Humans in Industrial COllaboration (\dataname{}, an excerpt in Fig.~\ref{fig:teaser}).
\dataname{} includes multi-view videos, 3D poses and trajectories of the joints of 20 human operators, in close collaboration with a robotic arm \emph{KUKA LBR iiwa} within a shared workspace. The dataset features 7 realistic industrial actions, taken at a real industrial assembly line with a marker-less setup.
The goal of \dataname{} is to endow cobots with perceptive awareness to enable human-cobot collaboration with contact. Towards this frontier, \dataname{} proposes to benchmark two key tasks: human pose forecasting and collision detection.
Cobots currently detect collisions by mechanical-only events (transmission of contact wrenches, control torques, sensitive skins). This ensures safety but it harms the human-cobot interaction, because collisions break the motion of both, which reduces productivity, and may be annoying to the human operator.
\dataname{} features 240 1-minute video recordings, from which two separate sets of test sequences are selected: one for estimating the accuracy in pose forecasting, so cobots may be aware of the next future (1.0 sec); and one with 226 genuine collisions, so cobots may foresee them and possibly re-plan.
The dataset is detailed in Sec.~\ref{sec:experimental_design}, experiments are illustrated in  Sec.~\ref{sec:experimental_results_sd}.

When tested on \dataname{}, the proposed \gcnname{} outperforms all baselines and reaches an error of 85.3 mm (MPJPE) at 1.00 sec, with a negligible run time of 2.3 msec (as reported in Table \ref{tab:collisions}). 
Additionally, the forecast human motion is used to detect human-cobot collision, by checking whether the predicted trajectory of the human body intersects that of the cobot. This is also encouraging, as \gcnname{} allows to reach an F1-score of 0.64.
Both aspects contribute to a cobot awareness of the future, which is instrumental for HRC in industrial applications.

\section{Related Work}
\label{sec:backgroung}

\smallskip
\myparagraph{Human pose forecasting.}
Human pose forecasting is a recent field which has some intersection with human action anticipation in computer vision~\cite{li20} and HRC~\cite{duarte2018action}. Previous studies exploited Temporal Convolutional Networks (TCNs)~\cite{Bai18TCN,Gehring17,Li18,pavllo19} and Recurrent Neural Networks (RNNs)~\cite{fragkiadaki15,anand19,jain16}. Both architectures are naturally suited to model the temporal dimension.
Recent works have expanded the range of available methods by using Variational Auto-Encoders~\cite{Btepage2017AnticipatingMF}, specific and model-agnostic layers that implicitly model the spatial structure of the human skeleton~\cite{Aksan_2019_ICCV}, or Transformer Networks~\cite{Cai20}.

\smallskip
\myparagraph{Pose forecasting using Graph Convolutional Networks (GCN).}
Most recent research uses GCNs~\cite{Dang21,li21,Mao20DCT,stsgcn,Zhao20}.
In \cite{Mao20DCT}, the authors  have mixed GCN for modelling the joint-joint interaction with Transformer Networks for the temporal patterns. Others~\cite{li21,stsgcn,Zhao20} have adopted GCNs to model the space-time body kinematics, devising, in the case of~\cite{Dang21}, hierarchical architectures to model coarse-to-fine body dynamics.\\
We identify three main research directions for improving efficiency in GCNs:
\textbf{i.}~space-time separable GCNs~\cite{stsgcn}, which factorizes the spatial joint-joint and temporal patterns of the adjacency matrix;
\textbf{ii.}~depth-wise separable graph convolutions~\cite{howard17}, which has been explored by~\cite{balcilar20} in the spectral domain;
and 
\textbf{iii.}~sparse GCNs~\cite{shi21}, which iteratively prunes the terms of the adjacency matrix of a GCN.
Notably, all three techniques also yield better performance than the plain GCN.
Here, for the first, we bring together these three aspects into an end-to-end space-time-depthwise-separable and sparse GCN. The three techniques are complementary to improve both efficiency and performance, but their integration requires some structural changes (\emph{e.g.},\ adopting teacher-student architectures for sparsifying), as we describe in Sec.~\ref{sec:methodology}.

\smallskip
\myparagraph{Human Robot Collaboration (HRC).}

HRC is the study of collaborative processes where human and robot agents work together to achieve shared goals~\cite{bauer2008human,castro2021trends}. Computer vision studies on HRC are mostly related to pose estimation ~\cite{OpenPoseTPAMI19,Garcia-Esteban2021,Lemmerz2019} to locate the articulated human body in the scene. 
 
 In~\cite{Chen2018,Kanazawa2019,Nascimento2020}, methodologies for robot motion planning and collision avoidance are proposed; their study perspective is opposite to ours, since we focus on the human operator. In this regard, the works of~\cite{Beltran2018,Costanzo2020,Kang2019,Lim2021} model the operators' whereabouts through detection algorithms which approximate human shapes using simple bounding boxes. 
Approaches that predict the human motion during collaborative tasks are in~\cite{torkar2019rnn,zhang2020recurrent} using RNNs and in~\cite{vianello2021human} using Guassian processes. Other work~\cite{laplaza2021attention} models the upper body and the human right hand (which they call the Human End Effector) by considering the robot-human handover phase. As motion prediction engine,  DCT-RNN-GCN~\cite{Mao20DCT} is considered, against which we compare in the experiments. 

\begin{table*}[t!]
\centering
\caption{Comparison between the state-of-the-art datasets and the proposed \dataname; \emph{unk} stands for ``unknown''. 
}
\resizebox{\textwidth}{!}{%
\begin{tabular}{l|ccccccc|c|cc|ccc|c|}
\cline{2-15}
\multirow{2}{*}{} &
  \multicolumn{7}{c|}{\textbf{Quantitative Details}} &
  \multirow{2}{*}{\textbf{\begin{tabular}[c]{@{}c@{}}Rec.\\ Scene\end{tabular}}} &
  \multicolumn{2}{c|}{\textbf{Actions Type}} &
  \multicolumn{3}{c|}{\textbf{Tasks}} &
  \multirow{2}{*}{\textbf{Markerless}}  \\ \cline{2-8} \cline{10-11} \cline{12-14} &
  \multicolumn{1}{c|}{\textbf{\begin{tabular}[c]{@{}c@{}}\#\\ Classes\end{tabular}}} &
  \multicolumn{1}{c|}{\textbf{\begin{tabular}[c]{@{}c@{}}\#\\ Subj.\end{tabular}}} &
  \multicolumn{1}{c|}{\textbf{\begin{tabular}[c]{@{}c@{}}Avg Rec.\\  Time\end{tabular}}} &
  \multicolumn{1}{c|}{\textbf{\begin{tabular}[c]{@{}c@{}}\#\\ Joints\end{tabular}}} &
  \multicolumn{1}{c|}{\textbf{FPS}} &
  \multicolumn{1}{c|}{\textbf{\begin{tabular}[c]{@{}c@{}}Aspect\\ Ratio\end{tabular}}} &
  \textbf{\begin{tabular}[c]{@{}c@{}}\# \\  Sensors\end{tabular}} &
   &
\multicolumn{1}{c|}{\textbf{Industr.}} &
\multicolumn{1}{c|}{\textbf{HRC}} &
\multicolumn{1}{c|}{\textbf{\begin{tabular}[c]{@{}c@{}}Action\\ Recog.\end{tabular}}} &
\multicolumn{1}{c|}{\textbf{\begin{tabular}[c]{@{}c@{}}Pose\\ Forec.\end{tabular}}} &
\multicolumn{1}{c|}{\textbf{\begin{tabular}[c]{@{}c@{}}Coll.\\ Det.\end{tabular}}} &
   \\ \hline

\multicolumn{1}{|l|}{Human3.6M ~\cite{h36m_pami}} &
  \multicolumn{1}{c|}{15} &
  \multicolumn{1}{c|}{11} &
  \multicolumn{1}{c|}{100.49 s} &
  \multicolumn{1}{c|}{32} &
  \multicolumn{1}{c|}{25} &
  \multicolumn{1}{c|}{normalized} &
  15 &
  \begin{tabular}[c]{@{}c@{}}mo-cap\\ studio\end{tabular} &
  \multicolumn{1}{c|}{} &
  \multicolumn{1}{c|}{} &
  \multicolumn{1}{c|}{} &
  \multicolumn{1}{c|}{\checkmark}
   & &\\ \hline

  \multicolumn{1}{|l|}{AMASS~\cite{AMASS:ICCV:2019}} &
  \multicolumn{1}{c|}{11265} &
  \multicolumn{1}{c|}{344} &
  \multicolumn{1}{c|}{12.89 s} &
  \multicolumn{1}{c|}{variable} &
  \multicolumn{1}{c|}{variable} &
  \multicolumn{1}{c|}{original} &
  variable &
  \begin{tabular}[c]{@{}c@{}}mo-cap\\ studio\end{tabular} &
 \multicolumn{1}{c|}{} &
  \multicolumn{1}{c|}{} &
  \multicolumn{1}{c|}{} &
  \multicolumn{1}{c|}{\checkmark}
   & &\\ \hline

\multicolumn{1}{|l|}{3DPW~\cite{vonMarcard2018}} &
  \multicolumn{1}{c|}{47} &
  \multicolumn{1}{c|}{7} &
  \multicolumn{1}{c|}{28.33 s} &
  \multicolumn{1}{c|}{18} &
  \multicolumn{1}{c|}{60} &
  \multicolumn{1}{c|}{original} &
  18 &
  \begin{tabular}[c]{@{}c@{}}outdoor\\ locations\end{tabular} &
 \multicolumn{1}{c|}{} &
  \multicolumn{1}{c|}{} &
  \multicolumn{1}{c|}{} &
  \multicolumn{1}{c|}{\checkmark}
   & &\\ \hline

\multicolumn{1}{|l|}{ExPI~\cite{guo2021multi}} &
  \multicolumn{1}{c|}{16} &
  \multicolumn{1}{c|}{4} &
  \multicolumn{1}{c|}{\emph{unk}} &
  \multicolumn{1}{c|}{18} &
  \multicolumn{1}{c|}{25} &
  \multicolumn{1}{c|}{original} &
  88 &
  \begin{tabular}[c]{@{}c@{}}mo-cap\\ studio\end{tabular} &
 \multicolumn{1}{c|}{} &
  \multicolumn{1}{c|}{} &
  \multicolumn{1}{c|}{} &
  \multicolumn{1}{c|}{\checkmark}
   & &\\ \hline

\multicolumn{1}{|l|}{CHI3D~\cite{fieraru2020three}} &
  \multicolumn{1}{c|}{8} &
  \multicolumn{1}{c|}{6} &
  \multicolumn{1}{c|}{\emph{unk}} &
  \multicolumn{1}{c|}{\emph{unk}} &
  \multicolumn{1}{c|}{\emph{unk}} &
  \multicolumn{1}{c|}{original} &
  14 &
  \begin{tabular}[c]{@{}c@{}}mo-cap \\ studio\end{tabular} &
 \multicolumn{1}{c|}{} &
  \multicolumn{1}{c|}{} &
  \multicolumn{1}{c|}{} &
  \multicolumn{1}{c|}{\checkmark}
   & &\\ \hline

\multicolumn{1}{|l|}{InHARD~\cite{Dallel2020}} &
  \multicolumn{1}{c|}{14} &
  \multicolumn{1}{c|}{16} &
  \multicolumn{1}{c|}{$<$ 8 s} &
  \multicolumn{1}{c|}{17} &
  \multicolumn{1}{c|}{120} &
  \multicolumn{1}{c|}{original} &
  20 &
  \begin{tabular}[c]{@{}c@{}}assembly \\ line\end{tabular} &
   \multicolumn{1}{c|}{\checkmark} &
  \multicolumn{1}{c|}{\checkmark} &
  \multicolumn{1}{c|}{\checkmark} &
  \multicolumn{1}{c|}{}
   & &\\ \hline    \hline
   %
 \multicolumn{1}{|l|}{\dataname (ours)} &
  \multicolumn{1}{c|}{7} &
  \multicolumn{1}{c|}{20} &
  \multicolumn{1}{c|}{55 s} &
  \multicolumn{1}{c|}{15} &
  \multicolumn{1}{c|}{25} &
  \multicolumn{1}{c|}{original} &
  3 &
  \begin{tabular}[c]{@{}c@{}}assembly \\ line\end{tabular} &
  \multicolumn{1}{c|}{\checkmark} &
  \multicolumn{1}{c|}{\checkmark} &
  \multicolumn{1}{c|}{} &
  \multicolumn{1}{c|}{\checkmark}
   &\checkmark &\checkmark\\ \hline    
\end{tabular}%
}

\label{tab:symbiotic_h36_comparative}
\end{table*}
 
\label{sec:cohre_dataset}
\smallskip
\myparagraph{Datasets for pose forecasting.}
Human pose forecasting datasets cover a wide spectrum of scenarios, see  Table~\ref{tab:symbiotic_h36_comparative} for a comparative analysis. 
Human3.6M~\cite{h36m_pami} considers everyday actions such as conversing, eating, greeting and smoking. Data were acquired using a 3D marker-based motion capture system, composed of 10 high-speed infrared cameras. AMASS~\cite{AMASS:ICCV:2019} is a collection of 15 datasets where daily actions were captured by an optical marker-based motion capture. Human3.6M and AMASS are standard benchmarks for human pose forecasting, with some overlap in the type of actions they deal with. The 3DPW dataset~\cite{vonMarcard2018} focuses on outdoor actions, captured with a moving camera and 17 Inertial Measurement Units (IMU), embedded on a special suit for motion capturing~\cite{Corrales2011}. The recent ExPI dataset~\cite{guo2021multi} contains 16 different dance actions performed by professional dancers, for a total of 115 sequences, and it is aimed at motion prediction. ExPI has been acquired with 68 synchronised and calibrated color cameras and a motion capture system with 20 mocap cameras. Finally, the CHI3D dataset~\cite{fieraru2020three} reports 3D data taken from MOCAP systems to study human interactions.


None of these datasets answer our research needs, i.e., a benchmark taken by a sparing, energy-efficient markerless system, focused on the industrial HRC scenario, where forecasting may be really useful for anticipating collisions between the humans and robots.
In fact, the only dataset relating to industrial applications is InHARD~\cite{Dallel2020}. Therein, humans are asked to perform an assembly task while wearing inertial sensors on each limb. The dataset is designed for human action recognition, and it involves 16 individuals performing 13 different actions each, for a total of 4800 action samples over more than 2 million frames. Despite showcasing a collaborative robot, in this dataset the robot is mostly static, making it unsuitable for collision forecasting. 


\section{Methodology}
\label{sec:methodology}
We 
build an accurate, memory efficient and fast GCN by bridging three diverse research directions: \textbf{i.}\ Space-time separable adjacency matrices; \textbf{ii.}\ Depth-wise separable graph convolutions; \textbf{iii.}\ Sparse adjacency matrices. This results in an all-Separable and Sparse GCN encoder for the human body kinematics, which we dub SeS-GCN, from which the future frames are forecast by a Temporal Convolutional Network (TCN).

\subsection{Background}

\myparagraph{Problem Formalization.} Pose forecasting is formulated as observing the 3D coordinates $\boldsymbol{x}_{v,t}$ of $V$ joints across $T$ frames and predicting their location in the $K$ future frames. For convenience of notation, we gather the coordinates from all joints at frame $t$ into the matrix $X_t=\left[ \boldsymbol{x}_{v,t} \right]_{v=1}^{V} \in \mathbb{R}^{3 \times V}$. Then we define the tensors $\mathcal{X}_{in}=[X_1,X_2...,X_T]$ and $\mathcal{X}_{out}=[X_{T+1},X_{T+2}...,X_{T+K}]$ that contain all observed input and target frames, respectively.

We consider a graph  $\mathcal{G}=(\mathcal{V},\mathcal{E})$ to encode the body kinematics, with all joints at all observed frames as the node set $\mathcal{V}=\left\{ \boldsymbol{v}_{i,t} \right\}_{i=1,t=1}^{V,T}$, and edges $(\boldsymbol{v}_{i,t},\boldsymbol{v}_{j,s})\in \mathcal{E}$ that connect joints $i, j$ at frames $t, s$.


\smallskip
\myparagraph{Graph Convolutional Networks (GCN).}
A GCN is a layered architecture:
\begin{equation}
\mathcal{X}^{(l+1)} = \sigma\left(A^{(l)} \mathcal{X}^{(l)} W^{(l)}\right)
\label{eq:gcn}
\end{equation}
The input to a GCN layer $l$ is the tensor $\mathcal{X}^{(l)} \in \mathbb{R}^{C^{(l)} \times V \times T}$ which maintains the correspondence to the $V$  body joints and the $T$ observed frames, but increases the depth of features to $C^{(l)}$ channels. $\mathcal{X}^{(1)}=\mathcal{X}_{in}$ is the input tensor at the first layer, with $C^{(1)}=3$ channels given by the 3D coordinates. $A^{(l)} \in \mathbb{R}^{VT\times VT}$ is the adjacency matrix relating pairs of $VT$ joints from all frames. Following most recent literature~\cite{Dang21,Mao20DCT,shi21,stsgcn}, $A^{(l)}$ is learnt.
$W^{(l)}\in \mathbb{R}^{C^{(l)} \times 1 \times 1}$ are the learnable weights of the graph convolutions. $\sigma$ is a the non-linear PReLU activation function.


\subsection{Separable \& Sparse Graph Convolutional Networks (SeS-GCN)}\label{sec:sepsparsegcn}

We build SeS-GCN by integrating the three mentioned modelling dimensions:
\textbf{i.}\ separating spatial and temporal interaction terms in the adjacency matrix of a GCN; \textbf{ii.}\ separating the graph convolutions depth-wise; \textbf{iii.}\ sparsifying the adjacency matrices of the GCN.


\smallskip
\myparagraph{Separating space-time.}
STS-GCN~\cite{stsgcn} has factored the adjacency matrix $A^{(l)}$ of the GCN, at each layer $l$, into the product of two terms $A^{(l)}_s\in\mathbb{R}^{V\times V\times T}$ and $A^{(l)}_t\in\mathbb{R}^{T\times T\times V}$, respectively responsible for the temporal-temporal and joint-joint relations. The GCN formulation becomes:
\begin{equation}
\mathcal{X}^{(l+1)} = \sigma\left(A^{(l)}_s A^{(l)}_t \mathcal{X}^{(l)} W^{(l)}\right)
\label{eq:stsgcn}
\end{equation}
Eq.~\eqref{eq:stsgcn} bottlenecks the interplay of joints across different frames, implicitly placing more emphasis on the interaction of joints on the same frame ($A^{(l)}_s$) and on the temporal pattern of each joint ($A^{(l)}_t$). This reduces the memory-footprint of a GCN by approx.\ 4x while improving its performance (cf.\ Sec.~\ref{subsec:model_choices}). Note that this differs from alternating spatial and temporal modules, as it is done in \cite{STAReccv20} and \cite{berta21}, respectively for trajectory forecasting and action recognition.

\smallskip
\myparagraph{Separating depth-wise.}
Inspired by depth-wise convolutions~\cite{chollet17,howard17}, 
the approach in~\cite{lai18} has introduced depth-wise graph convolutions for image classification, followed by~\cite{balcilar20} which resorted to a spectral formulation of depth-wise graph convolutions for graph classification. 
Here we consider depth-wise graph convolutions for pose forecasting. The depth-wise formulation bottlenecks the interplay of space and time (operated by the adjacency matrix $A^{(l)}$) with the channels of the graph convolution $W^{(l)}$. 
The resulting all-separable model which we dub STS-DW-GCN is formulated as such:
\begin{subequations}
\begin{align}
\mathcal{H}^{(l)} =&  \gamma\left(A^{(l)}_s A^{(l)}_t \mathcal{X}^{(l)} W_{DW}^{(l)}\right) \label{eq:allsep1}\\
\mathcal{X}^{(l+1)} &= \sigma\left(\mathcal{H}^{(l)} W_{\textnormal{MLP}}^{(l)}\right) \label{eq:allsep2}
\end{align}
\end{subequations}
Adding the depth-wise graph convolution splits the GCN of layer $l$ into two terms. The first, Eq.~\eqref{eq:allsep1}, focuses on space-time interaction and limits the channel cross-talk by the use of $W_{DW}^{(l)}\in \mathbb{R}^{\frac{C^{(l)}}{\alpha} \times 1 \times 1}$, with $1\le \alpha\le C^{(l)}$ setting the number of convolutional groups ($\alpha=C^{(l)}$ is the plain single-group depth-wise convolution). The second, Eq.~\eqref{eq:allsep2}, models the intra-channel communication just. This may be understood as a plain (MLP) 1D-convolution with $W_{\textnormal{MLP}}^{(l)}\in \mathbb{R}^{C^{(l)} \times 1 \times 1}$ which re-maps features from $C^{(l)}$ to $C^{(l+1)}$. $\gamma$ is the ReLU6 non-linear activation function. Overall, this does not significantly reduce the number of parameters, but it deepens the GCN without over-smoothing~\cite{Oono20}, which improves performance (see Sec.~\ref{subsec:model_choices} for details).

\smallskip
\myparagraph{Sparsifying the GCN.}
Sparsification has been used to improve the efficiency (memory
and, in some cases, runtime) of neural networks since the
seminal pruning work of \cite{braindamagLi89}.
\cite{shi21} has sparsified GCNs for trajectory forecasting.
This consists in learning masks $\mathcal{M}$ which selectively erase certain parameters in the adjacency matrix of the GCN.
Here we integrate sparsification with the all-separable GCN design, which yields our proposed SeS-GCN for human pose forecasting:
\begin{subequations}
\begin{align}
\mathcal{H}^{(l)} =& \gamma\left((\mathcal{M}^{(l)}_{s} \odot A^{(l)}_{s})( \mathcal{M}^{(l)}_{t}\odot A^{(l)}_{t}) \mathcal{X}^{(l)} W^{(l)}_{DW}\right) \label{eq:ses1}\\
\mathcal{X}^{(l+1)} &= \sigma\left(\mathcal{H}^{(l)} W_{\textnormal{MLP}}^{(l)}\right) \label{eq:ses2}
\end{align}
\end{subequations}
$\odot$ is the element-wise product and $\mathcal{M}^{(l)}_{\{s,t\}}$ are binary masks.
Both at training and inference, \cite{shi21} generates masks, it uses those to zero certain coefficients of the adjacency matrix $A$, and it adopts the resulting GCN for trajectory forecasting.
By contrast, we adopt a teacher-student framework during training. The teacher learns the masks, and the student only considers the spared coefficients in $A$.
At inference, our proposed SeS-GCN only consists of the student, which simply adopts the learnt sparse $A_{s}$ and $A_{t}$. Compared to \cite{shi21}, the approach of SeS-GCN is more robust at training, it yields fewer model parameters at inference 
($\sim$30\% less for both $A_{s}$ and $A_{t}$), and it reaches a better performance, as it is detailed in Sec.~\ref{subsec:model_choices}.

\subsection{Decoder Forecasting}

Given the space-time representation, as encoded by the SeS-GCN, the future frames are then decoded by using a temporal convolutional network (TCN) ~\cite{Bai18TCN,Gehring17,Li18,stsgcn}. The TCN remaps the temporal dimension to match the sought output number of predicted frames.
This part of the model is not considered for improvement because it is already efficient and it performs satisfactorily.

\section{The \dataname{} dataset}
\label{sec:experimental_design}

In this section the \dataname{} dataset is detailed by describing the acquisition scenario and devices, the cobot and the performed actions. We release RGB videos, skeletons and calibration parameters \footnote{Code and dataset are available at: \url{https://github.com/AlessioSam/CHICO-PoseForecasting}.}. 

\smallskip
\myparagraph{The scenario.}
We are in a smart-factory environment, with a single human operator standing in front of a $\SI{0.9}{m}\times\SI{0.6}{m}$ workbench and a cobot at its end (see Fig.\ref{fig:teaser}). The human operator has some free space to turn towards some equipment and carry out certain assembly, loading and unloading actions~\cite{michalos2015design}. In particular, light plastic pieces and heavy tiles, a hammer, abrasive sponges are available. The detailed setups for each action are reported graphically in the additional material. A total of 20 human operators have been hired for this study. They attended a course on how to operate with the cobot and signed an informed consent form prior to the recordings. 

\smallskip
\myparagraph{The collaborative robot.}
A 7 degrees-of-freedom Kuka LBR iiwa 14 R820 collaborates with the human operator during the data acquisition process. Weighing in at $\SI{29.5}{\kilogram}$ and with the ability to handle a payload up to $\SI{14}{\kilogram}$, it is widely used in modern production lines. More details on the cobot can be found in the supplementary material.

\smallskip
\myparagraph{The acquisition setup.}
The acquisition system is based on three RGB HD cameras 
providing three different viewpoints of the same workplace: two frontal-lateral, one rear view. 
The frame rate is 25Hz. 
The videos were first checked for erroneous or spurious frames, 
then we used Voxelpose~\cite{tu2020voxelpose} to extract 3D human pose for each frame.
Extrinsic parameters of each camera are estimated w.r.t. the robot's reference frame by means of a calibration chessboard of 1$\times$1m, and temporal alignment is guaranteed by synchronization of all the components with an Internet Time Server.
In our environment, Voxelpose estimates a joint positioning accuracy in terms of Mean Per Joint Position Error (MPJPE) of 24.99mm using three cameras, which is enough for our purposes, as an ideal compromise between portability of the system and accuracy. We confirm these numbers in two ways: the first is by checking that human-cobot collisions were detected with 100\% F1 score (we have a collision when the minimum distance between the human limbs and the robotic links is below a predefined threshold). Secondly, we show that the new CHICO dataset does not suffer from a trivial zero velocity solution~\cite{Julieta17}, \emph{i.e.}\ results achieved by a zero velocity model underperform the current SoA in equal proportion as for the large-scale established Human3.6M.

\smallskip
\myparagraph{Actions.}
The 7 types of actions of \dataname are inspired from ordinary work sessions in an HRC disassembly line as described in the review work of~\cite{HJORTH2022102208}. Each action is repeated over a time interval of $\sim$1 minute on average.
Each action is associated to a goal that the human operator has to achieve by a given time limit, which requires them to move with a certain velocity. Each action consists of repeated interactions with the robot (\emph{e.g.}, robot place, human picks) which, due to the limited space, lead to some \emph{unconstrained collisions}\footnote[1]{Unconstrained collisions is a term coming from~\cite{haddadin2009dlr}, indicating a situation
in which only the robot and human are directly involved into the collision.} which we label accordingly. 
Globally, from the 7 actions $\times$ 20 operators, we collect 226 different collisions. On the additional material, an excerpt of the videos with collisions are available. In the following, each action is shortly described.

\begin{itemize}
    \item \emph{\textbf{Lightweight pick and place}} (\emph{Light P\&P}). The human operator is required to move small objects of approximately 50 grams from a loading bay to a delivery location within a given time slot. The bay and the delivery location are at the opposite sides of the workbench. 
    Meanwhile, the robot loads on of this bay 
    so that the human operator has to pass close to the robotic arm.
    In many cases the distance between the limbs and the robotic arm is few centimeters. 
    \item \emph{\textbf{Heavyweight pick and place}} (\emph{Heavy P\&P}). The setup of this action is the same as before, but the objects to be moved are floor tiles weighing $\SI{0.75}{\kilogram}$. This means that the actions have to be carried out with two hands. 
    \item \emph{\textbf{Surface polishing}} (\emph{Polishing}). This action was inspired by~\cite{magrini2020human}, where the human operator polishes the border of a 40 by 60cm tile with some abrasive sponge, and the robot mimics a visual quality inspection. 
    \item \emph{\textbf{Precision pick and place}} (\emph{Prec. P\&P}). The robot places four plastic pieces in the four corners of a 30$\times$30cm table in the center of the workbench, and the human has to remove them and put on a bay, before the robot repeats the same unloading. 
    \item \emph{\textbf{Random pick and place}} (\emph{Rnd. P\&P}). Same as the previous action, except for the plastic pieces which were continuously placed by the robot randomly on the central 30$\times$30cm table, and the human operator has to remove them. 
    
    \item \emph{\textbf{High shelf lifting}} (\emph{High lift}). The goal was to pick light plastic pieces (50 grams each) on a sideway bay filled by the robot, putting them on a shelf located at 1.70m, at the opposite side of the workbench. Due to the geometry of the workspace, the arms of the human operator were required to pass above or below the moving robotic arm.
    In this way, close distances between the human arm and forearm and the robotic links were realized. 
    \item \emph{\textbf{Hammering}} (\emph{Hammer}). The operator hits with a hammer a metallic tide held by the robot. In this case, the interest was to check how much the collision detection is robust to an action where the human arm is colliding close to the robotic arm (that is, on the metallic tile) without properly colliding \emph{with the robotic arm}.
\end{itemize}



\section{Experiments on Human3.6M}
\label{sec:experimental_results_h36m}


We benchmark the proposed SeS-GCN model on the large and established Human3.6M~\cite{h36m_pami}. In Sec.~\ref{subsec:model_choices}, we analyze the design choices corresponding to the models discussed in Sec.~\ref{sec:methodology}, then we compare with the state-of-the-art in Sec.~\ref{subsec:comparison_h36m}. 

\medskip
\myparagraph{Human3.6M} \cite{h36m_pami} is an established dataset for pose forecasting, consisting of 15 daily life actions (e.g. Walking, Eating, Sitting Down). From the original skeleton of 32 joints, 22 are sampled as the task, representing the body kinematics. A total of 3.6 million poses are captured at 25 fps. In line with the literature~\cite{Mao20DCT,Julieta17,Dang21}, subjects 1, 6, 7, 8, 9 are used for for training, subject 11 for validation, and subject 5 for testing.

\begin{table}[b]
\centering
\caption{MPJPE error (millimeters) for long-range predictions (25 frames) on Human3.6M \cite{h36m_pami}  and numbers of parameters. Best figures overall are reported in bold, while underlined figures represent the best in each block.
The proposed model has comparable or less parameters than the GCN-based baselines~\cite{howard17,shi21,stsgcn} and it outperforms the best of them~\cite{stsgcn} by 2.6\%. 
%
}
\resizebox{1.\textwidth}{!}{
\begin{tabular}{l|c|c|c|c|c|c|c|c|}
\cline{2-9}
                                                 & Depth & MPJPE                             & Parameters (K)            & \multicolumn{1}{l|}{DW-Separable} & \multicolumn{1}{l|}{ST-Separable} & \multicolumn{1}{l|}{Sparse} & \multicolumn{1}{l|}{w/ MLP layers} & \multicolumn{1}{l|}{Teacher-Student} \\ \hline
\multicolumn{1}{|l|}{GCN}                        & 4     & 123.2                             & 222.7                        &                                   &                                   &                             &                                    &                                      \\ \hline
\multicolumn{1}{|l|}{DW-GCN \cite{lai18}}       & 4+4   & 119.8                            & 223.2                            & \checkmark         &                                   &                             &     \checkmark                               &                                      \\
\multicolumn{1}{|l|}{STS-GCN\footnote[2]{Results have been revisited. Please visit LINK} \cite{stsgcn} }     & 4     & \underline{117.0}                            & \textbf{\underline{57.6}}  &                                   & \checkmark         &                             &                                    &                                      \\
\multicolumn{1}{|l|}{Sparse-GCN \cite{shi21} } & 4     & 122.7                            & 257.9               &                                   &                                   & \checkmark   &                                    &                                      \\ \hline
\multicolumn{1}{|l|}{STS-GCN}              & 5     & 115.9                           & \underline{68.6}                        &                                   & \checkmark         &                             &                                    &                                      \\
\multicolumn{1}{|l|}{STS-GCN}              & 6     & 116.1                       & 79.9                              &                                   & \checkmark         &                             &                                    &                                      \\
\multicolumn{1}{|l|}{STS-GCN w/ MLP}               & 5+5   & 125.2                            & 101.4                            &                                   & \checkmark         &                             & \checkmark          &                                      \\
\multicolumn{1}{|l|}{STS-DW-GCN}              & 5+5   & \underline{114.8} & 70.0                              & \checkmark         & \checkmark         &                             & \checkmark          &                                      \\ \hline
\multicolumn{1}{|l|}{STS-DW-Sparse-GCN}    & 5+5   & 115.7                      & 122.4                             & \checkmark         & \checkmark         & \checkmark   & \checkmark          &                                      \\
\multicolumn{1}{|l|}{SeS-GCN (proposed)}         & 5+5   & \textbf{\underline{113.9}}   & \underline{58.6}    & \checkmark         & \checkmark         & \checkmark   & \checkmark          & \checkmark            \\ \hline
\end{tabular}}

\label{tab:ablation}
\end{table}

\medskip
\myparagraph{Metric.} \label{par:metrics} The prediction error is quantified via the MPJPE error metric~\cite{h36m_pami,Mao19LTD}, which considers the displacement of the predicted 3D coordinates w.r.t.\ the ground truth, in millimeters, at a specific future frame $t$:
%
\begin{equation}
{L_{\text{MPJPE}}} = {\frac{1}{V}\sum_{v=1}^{V}
||\hat{\boldsymbol{x}}_{vt}-\boldsymbol{x}_{vt} ||_{2}}.
\label{eqmpjpe}
\end{equation}
%

\subsection{Modelling choices of SeS-GCN}
\label{subsec:model_choices}
We review and quantify the impact of the modelling choices of SeS-GCN:

\medskip
\myparagraph{Efficient GCN baselines.}
In Table \ref{tab:ablation}, we first validate the three difference modelling approaches to efficient GCNs, namely space-time separable STS-GCN~\cite{stsgcn}, depth-wise separable graph convolutions DW-GCN \cite{lai18}, and Sparse-GCN~\cite{shi21}. STS-GCN yields the lowest MPJPE error of 117.0 mm at a 1 sec forecasting horizon (2.4\% better than DW-GCN, 4.8\% better than Sparse-GCN) with the fewest parameters, 57.6$k$ (ca.\ x4 less). We build therefore on this approach. 

\medskip
\myparagraph{Deeper GCNs.} 
It is a long standing belief that Deep Neural Networks (DNN) owe their performance to depth~\cite{resnet,vgg,resnext,zeilerfergus}. However, deeper models require more parameters and have a longer processing time. Additionally, deeper GCNs may suffer from over-smoothing~\cite{Oono20}. Seeking both better accuracy and efficiency, we consider three pathways for improvement:
(1) add GCN layers; (2) add MLP layers
between layers of GCNs; (3) adopt depth-wise graph convolutions, which also add MLP layers between GCN ones (cf.\ Sec.~\ref{sec:sepsparsegcn}).

As shown in Table \ref{tab:ablation}, there is a slight improvement in performance with 5 STS-GCN layers (MPJPE of 115.9 mm), but deeper models underperform. Adding MLP layers between the GCN ones (depth of 5+5) also decreases performance (MPJPE of 125.2). By contrast, adding depth by depth-wise separable graph convolutions (STS-DW-GCN of depth 5+5) reduces the error to 114.8 mm. This may be explained by the virtues of the increased depth in combination with the limiting cross-talk of joint-time-channels, which existing literature confirms~\cite{chollet17,lai18,stsgcn}. We note that space-time and depth-wise channel separability is complementary. Altogether, this performance is beyond the STS-GCN performance (114.8 Vs.\ 117.0 mm), at a slight increase of the parameter count (70$k$ Vs.\ 57.6$k$). 





\medskip
\myparagraph{Sparsifying GCNs and the proposed SeS-GCN.}
Finally, we target to improve efficiency by model compression.
Trends have reduced the size of models by reducing the parameter precision~\cite{rastegari2016xnornet}, by pruning and sparsifying some of the parameters~\cite{molchanov2017pruning}, or by constructing teacher-student frameworks, whereby a smaller student model is paired with a larger teacher to reach its same performance~\cite{hinton2015distilling,muyang20}. 
Note that the last technique is the current go-to choice in deploying very large networks such as Transformers~\cite{benesova2021costeffective}.

We start off by compressing the model with sparse adjacency matrices by the approach of Sparse-GCN~\cite{shi21}. They iteratively optimize the learnt parameters and the masks to select some (the selection occurs by a network branch, also at inference, cf.~\ref{sec:sepsparsegcn}).
As illustrated in Table~\ref{tab:ablation}, the approach of \cite{shi21} does not make a viable direction (STS-DW-Sparse-GCN), since the error increases to 115.7 mm and the parameter count to 122.4$k$.

Reminiscent of teacher-student models, in the proposed SeS-GCN we first train a teacher STS-DW-GCN, then use its learnt parameters to sparsify the affinity matrices of a student STS-DW-GCN, which is then trained from scratch. SeS-GCN achieves a competitive parameter count 
and the lowest MPJPE error of 113.9 mm, being comparable with the current SoA \cite{Mao20DCT} and using only 1.72\% of its parameters (58.6$k$ Vs. 3.4$M$).

\subsection{Comparison with the state-of-the-art (SoA)}
\label{subsec:comparison_h36m}
In Table \ref{tab:h36m_quantitative_full}, we evaluate the proposed SeS-GCN against three most recent techniques, over a short time horizon (10 frames, 400 msec) and a long time horizon (25 frames, 1000 msec). The first,
DCT-RNN-GCN~\cite{Mao20DCT}, the current SoA, uses DCT encoding, motion attention and RNNs and, differently from other models, demands more frames as input (50 vs. 10). The other two, MSR-GCN~\cite{Dang21} and STS-GCN~\cite{stsgcn} adopt GCN-only frameworks, the former adopts a multi-scale approach, the latter acts a separation between spatial and temporal encoding.

Both on Short- and long-term predictions, at the 400 and 1000 msec horizons, the proposed \gcnname{} outperforms other techniques~\cite{stsgcn,Mao20DCT} and it is within a 1.5\% error w.r.t.\ the current SoA~\cite{Mao20DCT}, while only using 1.72\% parameters and being $\sim$4 times faster than~\cite{Mao20DCT}.


\begin{savenotes}
\begin{table}[!htb]
\centering
\caption{
MPJPE error in mm for short-term (400 msec, 10 frames) and long-term (1000 msec, 25 frames) predictions of 3D joint positions on Human3.6M. The proposed model achieves competitive performance with the SoA~\cite{Mao20DCT}, while adopting 1.72\% of its parameters and running $\sim$4 times faster, cf.\ Table~\ref{tab:collisions}.
Results are discussed in  Sec.~\ref{subsec:comparison_h36m}.
}
\resizebox{\textwidth}{!}{
\begin{tabular}{c|cc|cc|cc|cc|cc|cc|cc|cc|}
\cline{2-17}
\multicolumn{1}{l|}{}               & \multicolumn{2}{c|}{Walking}             & \multicolumn{2}{c|}{Eating}              & \multicolumn{2}{c|}{Smoking}             & \multicolumn{2}{c|}{Discussion}          & \multicolumn{2}{c|}{Directions}          & \multicolumn{2}{c|}{Greeting}            & \multicolumn{2}{c|}{Phoning}             & \multicolumn{2}{c|}{Posing}              \\ \hline
\multicolumn{1}{|c|}{\textit{Time Horizon (msec)}} & {\ul \textit{400}} & {\ul \textit{1000}} & {\ul \textit{400}} & {\ul \textit{1000}} & {\ul \textit{400}} & {\ul \textit{1000}} & {\ul \textit{400}} & {\ul \textit{1000}} & {\ul \textit{400}} & {\ul \textit{1000}} & {\ul \textit{400}} & {\ul \textit{1000}} & {\ul \textit{400}} & {\ul \textit{1000}} & {\ul \textit{400}} & {\ul \textit{1000}} \\ \hline
\multicolumn{1}{|c|}{DCT-RNN-GCN \cite{Mao20DCT}}   & \textbf{39.8}              & \textbf{58.1}                & \textbf{36.2}               & \textbf{75.5}                & \textbf{36.4}               & \textbf{69.5}                & \textbf{65.4}               & 119.8               & 56.5               & 106.5               & \textbf{78.1}               & 138.8               & \textbf{49.2}               & 105.0               & \textbf{75.8}               & 178.2               \\
\multicolumn{1}{|c|}{MSR-GCN \cite{Dang21}}        & 45.2               & 63.0                & 40.4               & 77.1                & 38.1               & 71.6                & 69.7               & 117.5               & \textbf{53.8}               & \textbf{100.5}               & 93.3               & 147.2               & 51.2               & \textbf{104.3}               & 85.0               & 174.3               \\
\multicolumn{1}{|c|}{STS-GCN\footnote[2]{Results for STS-GCN differ from \cite{stsgcn}, due to revision by the authors, cf.\ \url{https://github.com/FraLuca/STSGCN}.
} \cite{stsgcn}}       & 51.0               & 70.2                & 43.3               & 82.6                & 42.3               & 76.1                & 71.9               & 118.9                & 63.2              & 109.6                & 86.4               &  \textbf{136.1}                & 53.8               & 108.3                & 84.7               & 178.4               \\ \hline
\multicolumn{1}{|c|}{SeS-GCN (proposed)}         
& 48.8     & 67.3       & 41.7      & 78.1       & 40.8      & 73.7       & 70.6      & \textbf{116.7}       & 60.3      & 106.9       & 83.8      & 137.2       & 52.6      & 106.7       & 82.6      & \textbf{173.5}      \\ \hline
\end{tabular}}
\end{table}
\end{savenotes}


\begin{table}[!htb]
\centering

\resizebox{\textwidth}{!}{
\begin{tabular}{c|cc|cc|cc|cc|cc|cc|cc|cc|}
\cline{2-17}
\multicolumn{1}{l|}{\textbf{}}      & \multicolumn{2}{c|}{Purchases}           & \multicolumn{2}{c|}{Sitting}             & \multicolumn{2}{c|}{Sitting Down}        & \multicolumn{2}{c|}{Taking Photo}        & \multicolumn{2}{c|}{Waiting}             & \multicolumn{2}{c|}{Walking Dog}         & \multicolumn{2}{c|}{Walking Together}    & \multicolumn{2}{c|}{Average}             \\ \hline
\multicolumn{1}{|c|}{\textit{Time Horizon (msec)}} & {\ul \textit{400}} & {\ul \textit{1000}} & {\ul \textit{400}} & {\ul \textit{1000}} & {\ul \textit{400}} & {\ul \textit{1000}} & {\ul \textit{400}} & {\ul \textit{1000}} & {\ul \textit{400}} & {\ul \textit{1000}} & {\ul \textit{400}} & {\ul \textit{1000}} & {\ul \textit{400}} & {\ul \textit{1000}} & {\ul \textit{400}} & {\ul \textit{1000}} \\ \hline
\multicolumn{1}{|c|}{DCT-RNN-GCN \cite{Mao20DCT}}   & \textbf{73.9}               & \textbf{134.2}               & \textbf{56.0}               & \textbf{115.9}               & \textbf{72.0}               & \textbf{143.6}               & \textbf{51.5}               & \textbf{115.9}               & \textbf{54.9}               & 108.2               & \textbf{86.3}               & \textbf{146.9}               & \textbf{41.9}               & \textbf{64.9}                & \textbf{58.3}               & \textbf{112.1}               \\
\multicolumn{1}{|c|}{MSR-GCN \cite{Dang21}}        & 79.6               & 139.1               & 57.8               & 120.0               & 76.8               & 155.4               & 56.3               & 121.8               & 59.2               & \textbf{106.2}               & 93.3               & 148.2               & 43.8               & 65.9                & 62.9               & 114.1               \\
\multicolumn{1}{|c|}{STS-GCN\footnote[2]{} \cite{stsgcn}}       & 83.1               & 141.0                & 60.8               & 121.4                & 79.4               & 148.4                & 59.4               & 126.3                & 62.0               & 113.6                & 97.3               & 151.5               & 49.1               & 72.5       & 65.8               & 117.0                \\ \hline
\multicolumn{1}{|c|}{SeS-GCN (proposed)}        
& 82.2      & 139.1       & 59.9      & 117.5       & 78.1      & 146.0       & 57.7     & 121.2       & 58.5      & 107.5       & 94.0      & 147.7      & 48.3      & 70.8                & 64.0      & 113.9       \\ \hline
\end{tabular}}

\label{tab:h36m_quantitative_full}

\end{table}

\section{Experiments on \dataname{}}
\label{sec:experimental_results_sd}

We benchmark on \dataname{} the SoA and the proposed \gcnname{} model. The two HRC tasks of human pose forecasting and collision detection are discussed in Secs.~\ref{subsec:quantitative_symb} and \ref{sec:coll} respectively.

\subsection{Pose forecasting benchmark}
\label{subsec:quantitative_symb}

Here we describe the evaluation protocol proposed for \dataname{} and report comparative evaluation of pose forecasting techniques.

\myparagraph{Evaluation protocol.}
We create the train/validation/test split by assigning 2 subjects to the validation (subjects 0 and 4), 4 to the test set (subjects 2, 3, 18 and 19), and the remaining 14 to the training set.
For short-range prediction experiments, abiding the setup of Human3.6M~\cite{h36m_pami}, we consider 10 frames as observation time and 10 or 25 frames as forecasting horizon.
Differently from all reported techniques, DCT-RNN-GCN requires 50 input frames.

\noindent
We adopt the same Mean Per Joint Position Error (MPJPE)\cite{h36m_pami}
as Human3.6M, in Eq.~\eqref{eqmpjpe}, which also defines the training loss for the evaluated techniques.

\noindent
None of the motion sequences for pose forecasting contains collisions. In fact, the objective is to train and test the ``correct'' collaborative human behavior, and not the human retractions and the pauses due to the collisions\footnote[7]{After the collisions, the robot stops for 1 seconds, during which the human operator usually stands still, waiting for the robot to resume operations.}.

\begin{table}[!h]
\centering
\caption{ MPJPE error in mm for short-term (400 msec, 10 frames) and long-term (1000 msec, 25 frames) prediction of 3D joint positions on \dataname dataset. 
The average error is 7.9\% lower than the other models in the short-term and 2.4\% lower in the long-term prediction. See Sec. \ref{subsec:quantitative_symb} for a discussion.}
\resizebox{1.\textwidth}{!}{
\begin{tabular}{c|cc|cc|cc|cc|cc|cc|cc|cc|}
\cline{2-17}
\multicolumn{1}{l|}{}                              & \multicolumn{2}{c|}{Hammer}              & \multicolumn{2}{c|}{High Lift}           & \multicolumn{2}{c|}{Prec. P\&P}          & \multicolumn{2}{c|}{Rnd. P\&P}                     & \multicolumn{2}{c|}{Polishing}           & \multicolumn{2}{c|}{Heavy P\&P}          & \multicolumn{2}{c|}{Light P\&P}          & \multicolumn{2}{c|}{Average}             \\ \hline
\multicolumn{1}{|c|}{\textit{Time Horizon (msec)}} & {\ul \textit{400}} & {\ul \textit{1000}} & {\ul \textit{400}} & {\ul \textit{1000}} & {\ul \textit{400}} & {\ul \textit{1000}} & {\ul \textit{400}}           & {\ul \textit{1000}} & {\ul \textit{400}} & {\ul \textit{1000}} & {\ul \textit{400}} & {\ul \textit{1000}} & {\ul \textit{400}} & {\ul \textit{1000}} & {\ul \textit{400}} & {\ul \textit{1000}}  \\ \hline
\multicolumn{1}{|c|}{DCT-RNN-GCN \cite{Mao20DCT}}                  & 41.1              & \textbf{39.0}                & 69.4               & 128.8               & 50.6               & 83.3                & 52.7                         & 88.2                & 42.1               & 76.0                & 64.1               & 121.5               & 62.1               & 104.2                & 54.6        & 91.6       \\
\multicolumn{1}{|c|}{MSR-GCN \cite{Dang21}}                       & 41.6      & 39.7       & 67.8               & 130.2               & 50.2               & 81.3                & 53.4 & 90.3                & 41.1      & 73.2       & 62.7               & 118.2               & 61.5               & 101.9               & 54.1               & 90.7    \\
\multicolumn{1}{|c|}{STS-GCN \cite{stsgcn}}                      & 46.6               & 52.1                & 64.2              & 116.4               & 48.3               & 79.5                & 52.0                         & 87.9                & 42.1               & 73.9                & 60.6               & 106.5       & 57.2               & 95.2                & 53.0               & 87.4                \\ \hline
\multicolumn{1}{|c|}{SeS-GCN (proposed)}           & \textbf{40.9}               & 49.3                & \textbf{62.1}      & \textbf{116.3}      & \textbf{46.0}      & \textbf{77.4}       & \textbf{48.4}                & \textbf{84.8}       & \textbf{38.8}               & \textbf{72.4}                & \textbf{56.1}      & \textbf{104.4}                & \textbf{56.2}      & \textbf{92.2}       & \textbf{48.8}      & \textbf{85.3}      \\ \hline
\end{tabular}
}

\label{tab:CHICOresults}
\end{table}



\myparagraph{Comparative evaluation.}
In Table~\ref{tab:CHICOresults}, we compare pose forecasting techniques from the SoA and the proposed \gcnname{}. On the short-term predictions the best performance is that of \gcnname{}, reaching an MPJPE error of 48.8 mm, which is 7.9\% better than the second best STS-GCN~\cite{stsgcn}.

\noindent
On the longer-term predictions, the best performance (MPJPE error of 85.3 mm) is also detained by \gcnname{}, which is 2.4\% better than the second best STS-GCN~\cite{stsgcn}. 
The proposed model outperforms all techniques on all actions except \emph{Hammer}, a briefly repeating action which may differ for single hits.
We argue that DCT-RNN-GCN~\cite{Mao20DCT} may get an advantage from using 50 input frames (all other methods use 10 frames)

\begin{figure}[!b]
    \centering

    \includegraphics[width=0.5\linewidth]{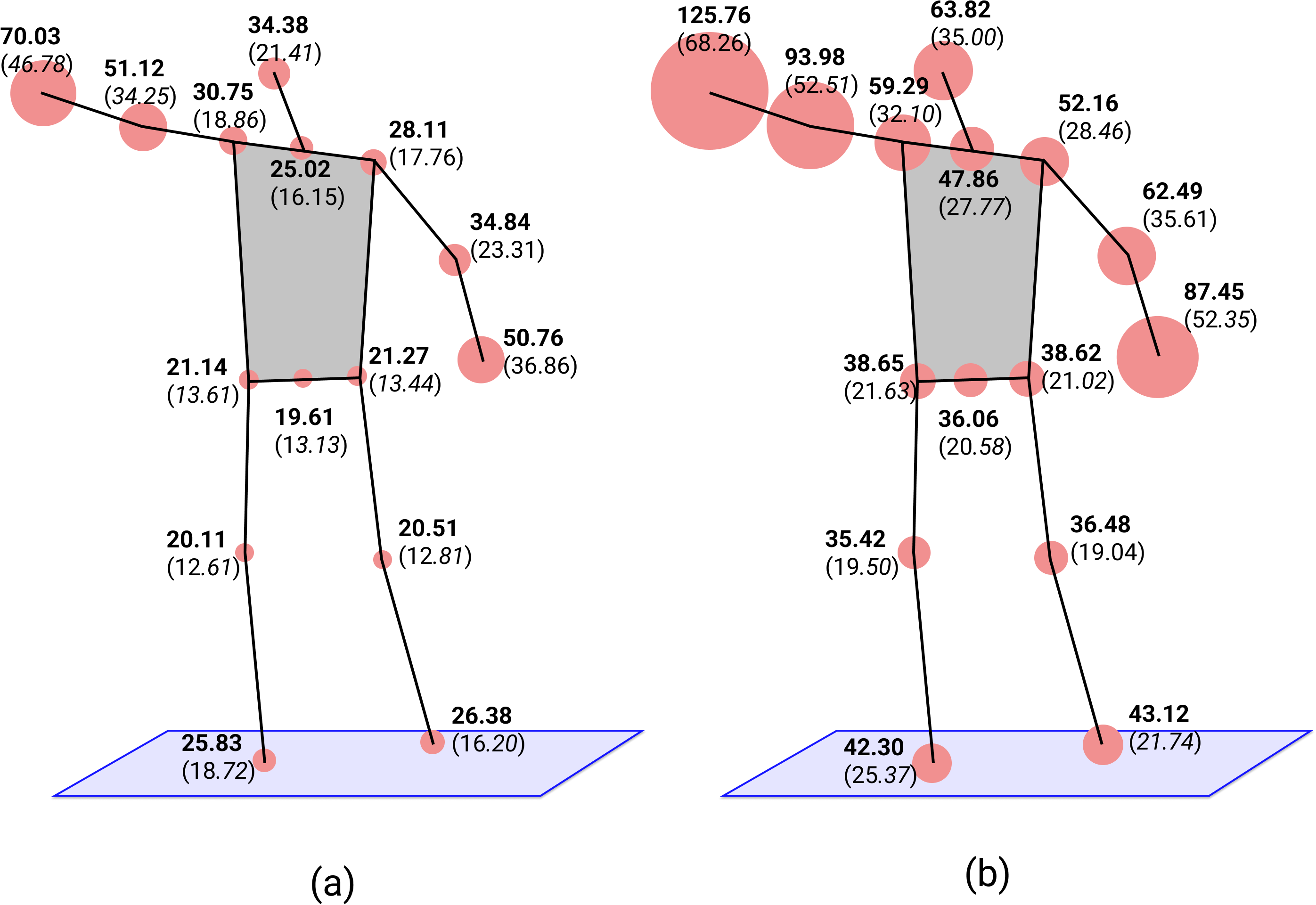}
    \caption{Average MPJPE distribution for all actions in \dataname on different joints for (a) short-term (0.40 s) and (b) long-term (1.00 s) predictions. The radius of the blob gives the spatial error with the same scale of the skeleton.} 
    \label{fig:error}
\end{figure}

For a graphical illustration, Fig.~\ref{fig:error} shows a distribution of the error per joint calculated over all the actions, for the horizons 400 (\textit{left}) and 1000 msec (\textit{right}). In both cases the error gets larger as we get closer to the extrema of the kinematic skeleton, since those joints move the most.
The slightly larger error at the right hands (70.03 and 125.76 mm, respectively) matches that subjects are right-handed (but some actions are operated with both hands).

For a sanity check of results, we have also evaluated the performance of a trivial zero velocity model. \cite{Julieta17} has found that keeping the last observed positions may be a surprisingly strong (trivial) baseline. For CHICO, the zero velocity model scores an MPJPE of 110.6 at 25-frames, worse than the 85.3 mm score of SeS-GCN. This is in line with the large-scale dataset Human3.6M~\cite{h36m_pami}, where the performance of the trivial model is 153.3 mm.

\subsection{Collision detection experiments}\label{sec:coll}


\myparagraph{Evaluation protocol.}
We consider a collision to occur when any body limb of the subject gets too close to any part of the cobot, i.e.\ within a distance threshold, for at least one frame. In particular, a collision refers to the proximity between the cobot and the human in the forecast portion of the trajectory. The (Euclidean) distance threshold is set to 13 cm.

\noindent
The motion of the cobot is scripted beforehand, thus known.
The motion of the human subjects in the next 1000 msec needs to be forecast, starting from the observation of 400 msec. The train/validation/test sets sample sequences of 10+25 frames with stride of 10.

\myparagraph{Evaluation of collision detection.}
For the evaluation of collision, following \cite{minelli2020iros}, both the cobot arm parts and the human body limbs are approximated by cylinders. The diameters for the cobot are fixed to 8cm. Those of the body limbs are taken from a human atlas.

\noindent
In Table~\ref{tab:collisions}, we report precision, recall and F$_1$ scores for the detection of collisions on the motion of 2 test subjects, which contains 21 collisions.
The top performer in pose forecasting, our proposed \gcnname{}, also yields the largest F$_1$ score of 0.64. The lower performing MSR-GCN~\cite{Dang21} yields poor collision detection capabilities, with an F$_1$ score of 0.31.

\noindent
\begin{table}[!h]
      \centering
    \caption{Evaluation of collision detection performance achieved by competing pose forecasting techniques, with indication of inference run time. See discussion in Sec.~\ref{sec:coll}.}
      \resizebox{6cm}{!}{
        \begin{tabular}{|l|c|c|c||c|}
          \cline{1-4}
          \textit{Time Horizon (msec)} & \multicolumn{3}{c||}{ \textbf{\textit{1000}}} \\ \hline
          \textit{Metrics} &  \textit{Prec} & \textit{Recall} & \textit{F}$_1$ & \textit{Inference Time (sec)} \\ \hline
          \multicolumn{1}{|c|}{DCT-RNN-GCN~\cite{Mao20DCT}}& 0.63 & 0.58 & 0.56 &  $9.1 \times 10^{-3}$\\
          \multicolumn{1}{|c|}{MSR-GCN~\cite{Dang21}}& 0.63 & 0.30 & 0.31 & $25.2 \times 10^{-3}$\\
          \multicolumn{1}{|c|}{STS-GCN~\cite{stsgcn}}& 0.68 & 0.61 & 0.63 & {$\mathbf{2.3 \times 10^{-3}}$} \\ \hline
          \multicolumn{1}{|c|}{SeS-GCN (proposed)} & 0.84 & 0.54 & \textbf{0.64} & {$\mathbf{2.3 \times 10^{-3}}$}\\ \hline        \end{tabular}
      }

    \label{tab:collisions}  
\end{table}
\noindent
\section{Conclusions}
\label{sec:conclusion}
Towards the goal of forecasting  the human motion during human-robot collaboration in industrial (HRC) environments, we have proposed the novel \gcnname{} model, which integrates three most recent modelling methodologies for accuracy and efficiency: space-time separable GCNs, depth-wise separable graph convolutions and sparse GCNs. 
Also, we have contributed a new \dataname{} dataset, acquired at real assembly line, the first providing a benchmark of the two fundamental HRC tasks of human pose forecasting and collision detection.
Featuring an MPJPE error of 85.3 mm at 1 sec in the future with a negligible run time of 2.3 msec, \gcnname{} and \dataname{} unleash great potential for perception algorithms and their application in robotics.
\\
\\
\myparagraph{Acknowledgements.} This work was supported by the Italian MIUR through the project "Dipartimenti di Eccellenza 2018-2022", and partially funded by DsTech S.r.l.
\clearpage

\section{Appendix}
We supplement the main paper submission with extra information in this section, as follows:
\begin{itemize}
    \item Sec.~\ref{sec:chico_viz} supplements the main paper with more detailed descriptions and illustrations of each action in \dataname. This extends the descriptions and illustrations in the \textit{main paper}.
    \item Sec.~\ref{sec:impl_det} discusses some additional details regarding the implementation and learning of SeS-GCN.
\end{itemize}

\subsection{The CHICO Dataset}
\label{sec:chico_viz}

Here we illustrate additional details on the scenario of the dataset and the acquisition process; furthermore,  we report a graphical explanation of all the 8 actions.
For further details, please watch the additional video, enclosed in \url{https://github.com/AlessioSam/CHICO-PoseForecasting}.

\subsubsection{Details on the dataset and the data acquisition process.}
The dataset has been acquired in the October '21 - March '22 period, on a $500m^2$ Industry 4.0 lab, which includes a configurable a 11m production line, 4 cobots, a quality inspection cell, a (dis-)assembly station and other equipment. We worked on the 0.9 m ×0.6 m workbench of the (dis-)assembly station in front of a  Kuka LBR iiwa 14 R820 cobot. The declared positioning accuracy is $\pm \SI{0.1}{mm}$ and the axis-specific torque accuracy is $\pm \SI{2}{\%}$~\cite{kuka_manual}. Thanks to its joint torque sensors, the robot can detect contact and reduce its level of force and speed, being compliant to the ISO/TS 15066:2016~\cite{iso15066} standard. Since collisions between the operator and the cobot were expected in \dataname{}, the maximum allowed Cartesian speed of each link is set to $\SI{200}{mm/s}$, slightly lower than the ISO/TS 15066:2016 requirements. The safety torque limit allowed before the mechanical brakes activation is set to $\SI{30}{\newton\meter}$ for all joints and $\SI{50}{\newton\meter}$ for the end-effector. Additionally, a programmable safety check of $\SI{10}{\newton}$ was set on the Cartesian force. 

A total of 20 subjects (14 males, 6 females, average age 23 years) have been hired for building the dataset. They worked for the entire acquisition period, after having signed an informed consent and participated on a crash course on how to cooperate with the Kuka cobot. During the acquisition season, we have selected some excerpts which capture collisions made by the operators, reaching 226 collisions.
In average we have 45 collisions for each action, with the sole exception for \textit{hammering}. For this specific action, the cobot stands still, holding the object being hammered, while the human agent moves repeatedly close to the robotic arm.
Sequences containing this action are still part of the collision detection dataset, i.e.\ they are useful to check that there are no false positives.

\subsubsection{Details on the actions.}
In this section we expand the short explanations of each action given in Sec.4 of the main paper. For each action, we report the original description, and a detailed graphical storyboard.

\begin{itemize}
    \item \emph{\textbf{Lightweight pick and place}} (\emph{Light P\&P}). The human operator is required to move small objects of approximately 50 grams from a loading bay to a delivery location within a given time slot. The bay and the delivery location are at the opposite sides of the workbench. 
    Meanwhile, the robot loads on of this bay 
    so that the human operator has to pass close to the robotic arm.
    In many cases the distance between the limbs and the robotic arm is few centimeters. 

 \begin{figure}[h]
    \centering
    \includegraphics[width=1.0\linewidth]{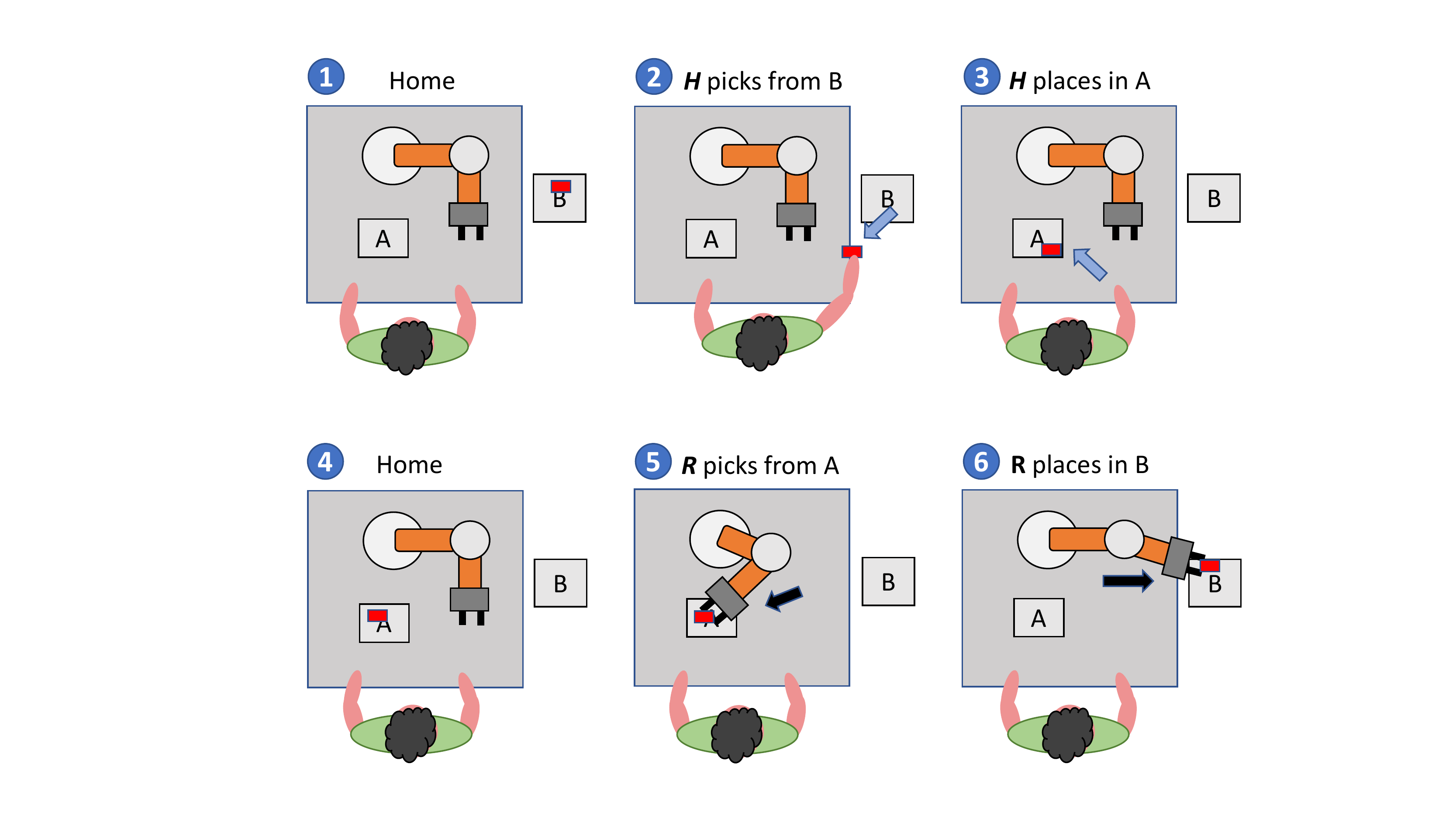}
    \caption{\emph{\textbf{Lightweight pick and place}} illustration; ``\emph{\textbf{H}}'' stands for ``human operator'', ``\emph{\textbf{R}}'' for ``robot''.  A single item (the red brick) is showed here for clarity. In practice, a dozen of items was available.
    }
    \label{fig:span_light}
\end{figure}
\end{itemize}
\newpage
\begin{itemize}
    \item \emph{\textbf{Heavyweight pick and place}} (\emph{Heavy P\&P}). The setup of this action is the same as before, but the objects to be moved are floor tiles weighing $\SI{0.75}{\kilogram}$.
    This means that the actions have to be carried out with two hands. 
\end{itemize}
 \begin{figure}[h]
    \centering
  \includegraphics[width=1.0\linewidth]{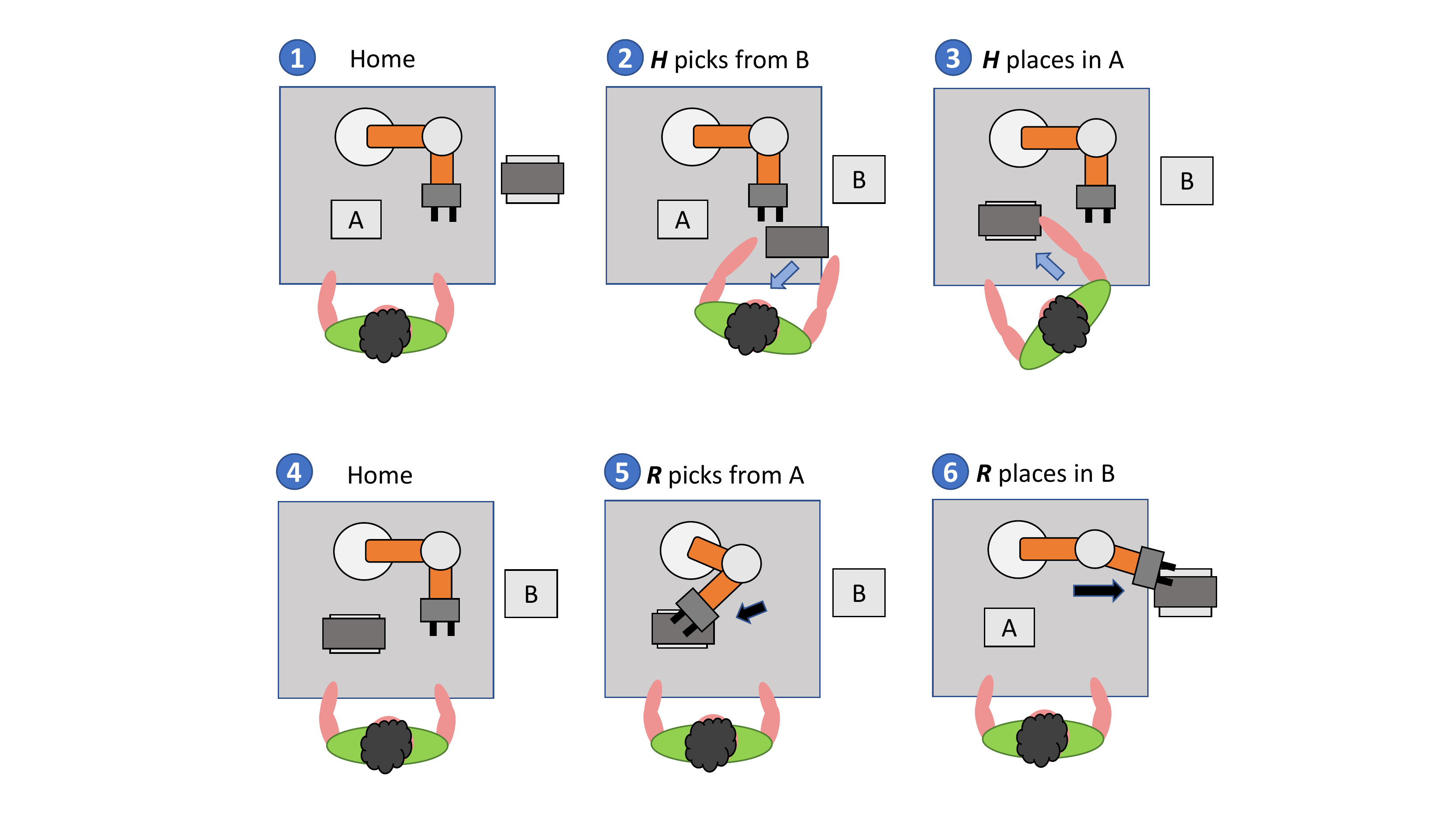}
    \caption{\emph{\textbf{Heavyweight pick and place}} illustration; ``\emph{\textbf{H}}'' stands for ``human operator'', ``\emph{\textbf{R}}'' for ``robot''.
    Moving the object with two hands requires to strongly rotating the torso, and this partially hides the robot from the operator, being back to him/her. This was the main cause of the collisions occurred with this action.}
    \label{fig:span_light}
\end{figure}
\newpage
\begin{itemize}
     \item \emph{\textbf{Surface polishing}} (\emph{Polishing}). This action was inspired by~\cite{magrini2020human}, where the human operator polishes the border of a 40 by 60cm tile with some abrasive sponge, and the robot mimics a visual quality inspection.
     
\end{itemize}
 \begin{figure}[h]
    \centering
  \includegraphics[width=1.0\linewidth]{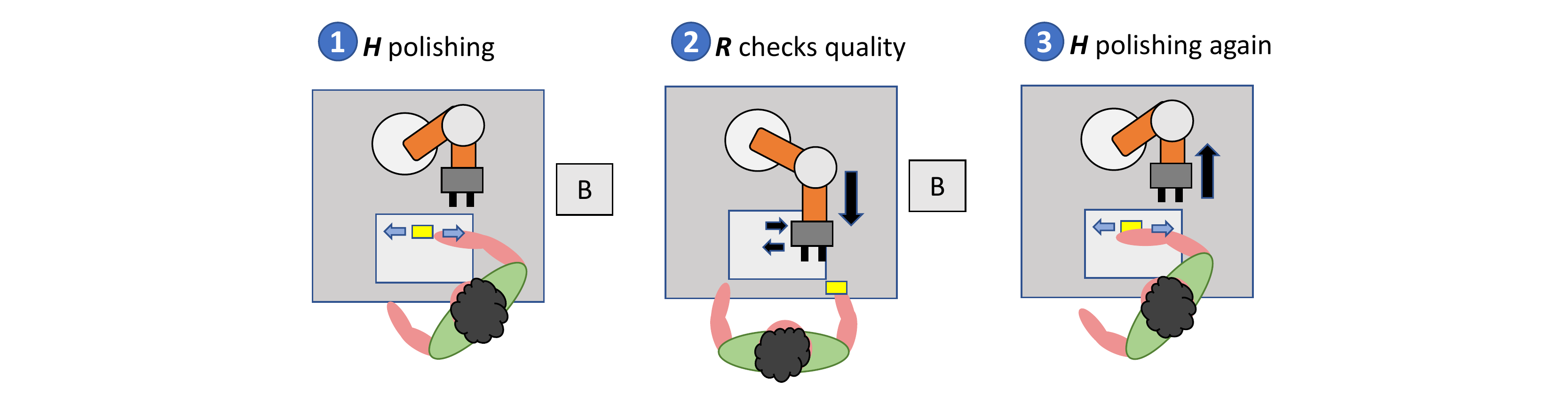}
    \caption{ \emph{\textbf{Surface polishing}} illustration; ``\emph{\textbf{H}}'' stands for ``human operator'', ``\emph{\textbf{R}}'' for ``robot''. The human has an abrasive sponge used to remove some material from the metallic tile. This action created most of the collisions,  since the action required the user to be prone on the surface to polish, blocking the view of the robot.}
    \label{fig:span_light}
\end{figure}
\newpage

\begin{itemize}
     \item \emph{\textbf{Precision pick and place}} (\emph{Prec. P\&P}). The robot places four plastic pieces in the four corners of a 30$\times$30cm table in the center of the workbench, and the human has to remove them and put on a bay, before the robot repeats the same unloading. 
\end{itemize}
 \begin{figure}[h]
    \centering
  \includegraphics[width=1.0\linewidth]{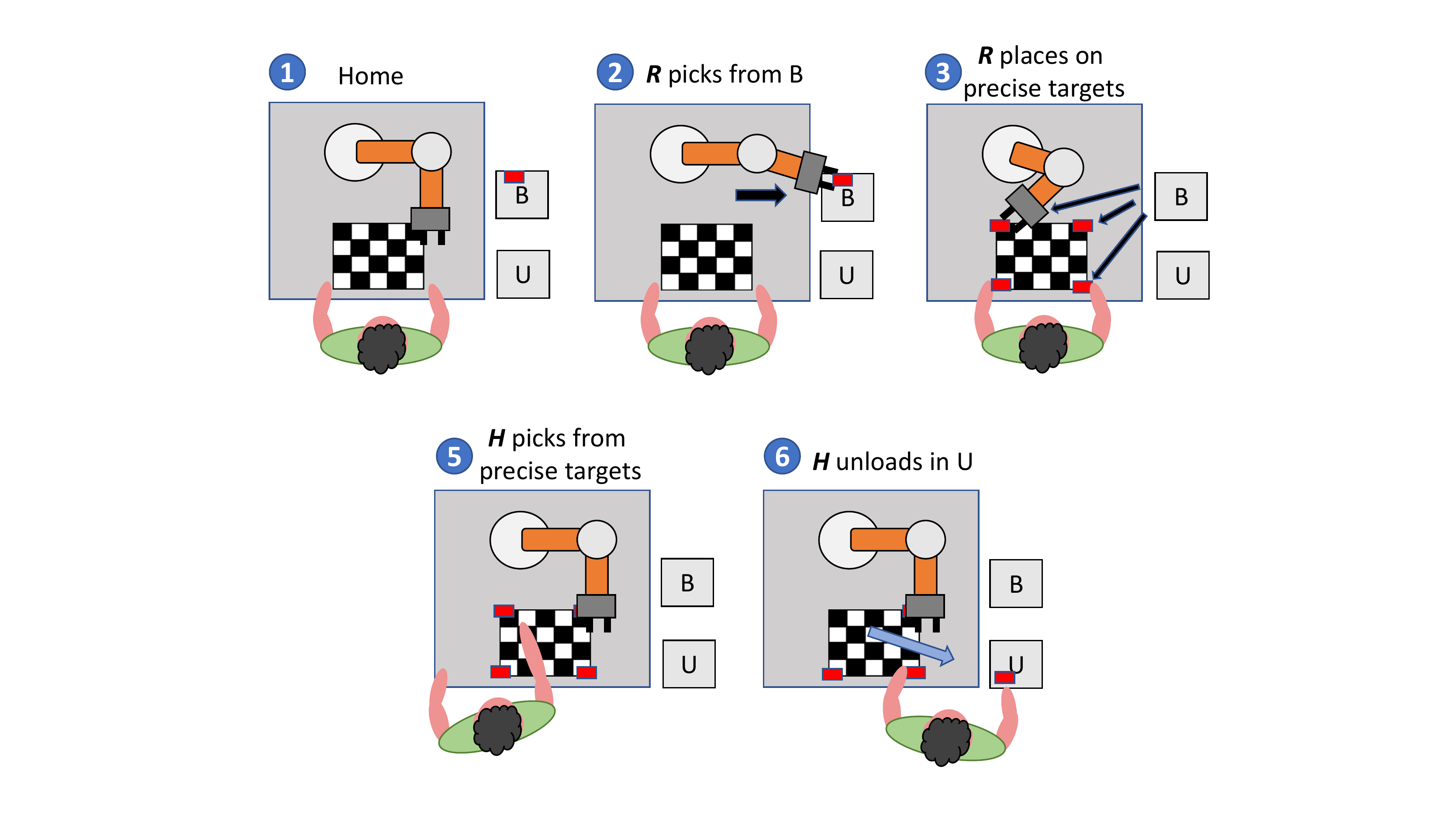}
    \caption{\emph{\textbf{Precision pick and place}} illustration; ``\emph{\textbf{H}}'' stands for ``human operator'', ``\emph{\textbf{R}}'' for ``robot''. This action is important, since it allows to measure how precise the prediction could be in individuating particular endpoints which will be targeted by the human operator.}
    \label{fig:span_light}
\end{figure}
\newpage
\begin{itemize}
    \item \emph{\textbf{Random pick and place}} (\emph{Rnd. P\&P}). Same as the previous action, except for the plastic pieces which were continuously placed by the robot randomly on the central 30$\times$30cm table, and the human operator has to remove them. 
\end{itemize}
 \begin{figure}[h]
    \centering
  \includegraphics[width=1.0\linewidth]{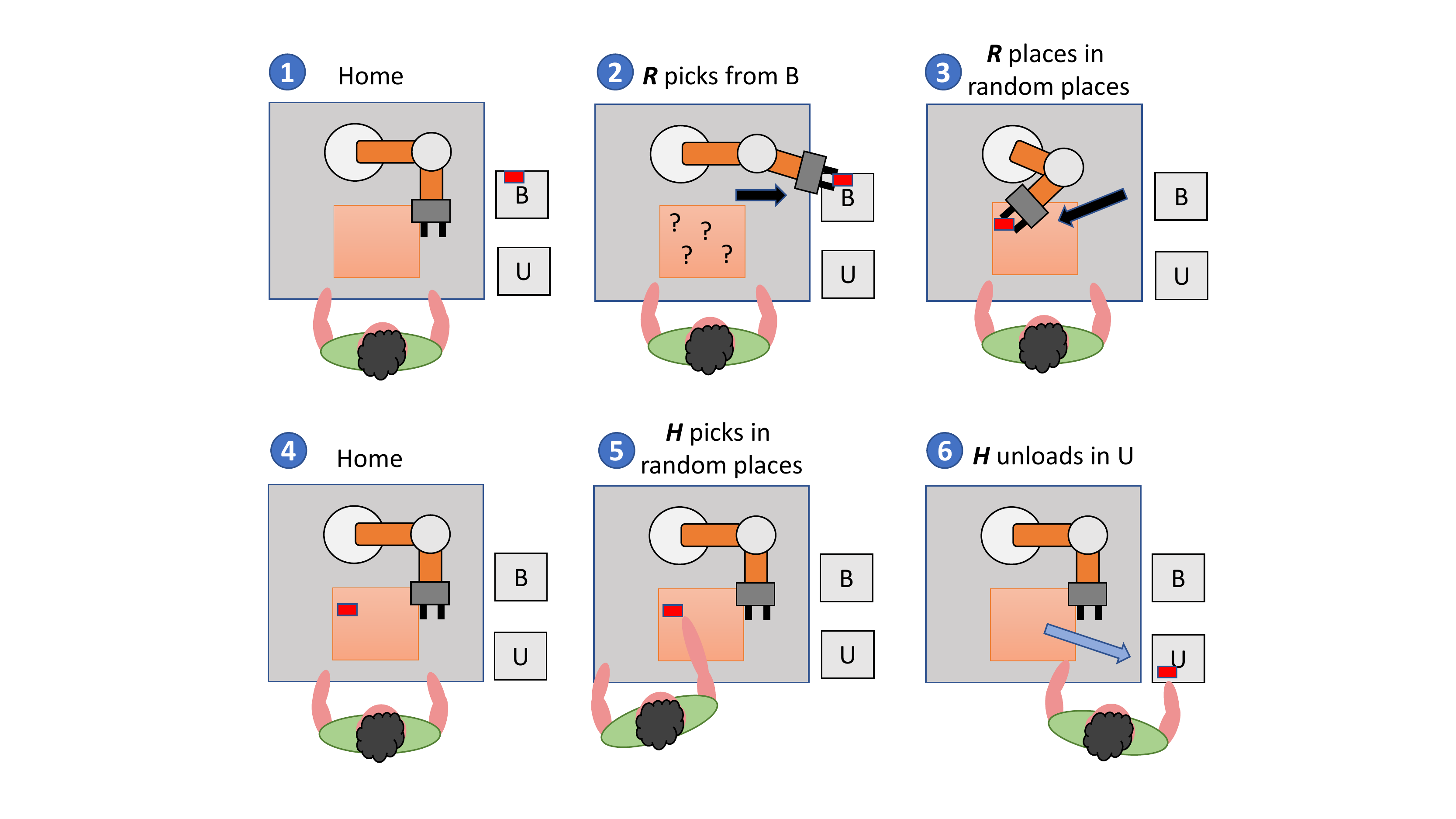}
    \caption{\emph{\textbf{Random pick and place}} illustration; ``\emph{\textbf{H}}'' stands for ``human operator'', ``\emph{\textbf{R}}'' for ``robot''. The action is interesting, since the robot puts objects randomly on the workplace, and this created some collisions during the interaction. A single item (the red brick) is showed here for clarity. In practice, a dozen of items was available.}
    \label{fig:span_light}
\end{figure}
\newpage
\begin{itemize}
   \item \emph{\textbf{High shelf lifting}} (\emph{High lift}). The goal was to pick light plastic pieces (50 grams each) on a sideway bay filled by the robot, putting them on a shelf located at 1.70m, at the opposite side of the workbench. Due to the geometry of the workspace, the arms of the human operator were required to pass above or below the moving robotic arm.    In this way, close distances between the human arm and forearm and the robotic links were realized. 
\end{itemize}
 \begin{figure}[h]
    \centering
  \includegraphics[width=1.0\linewidth]{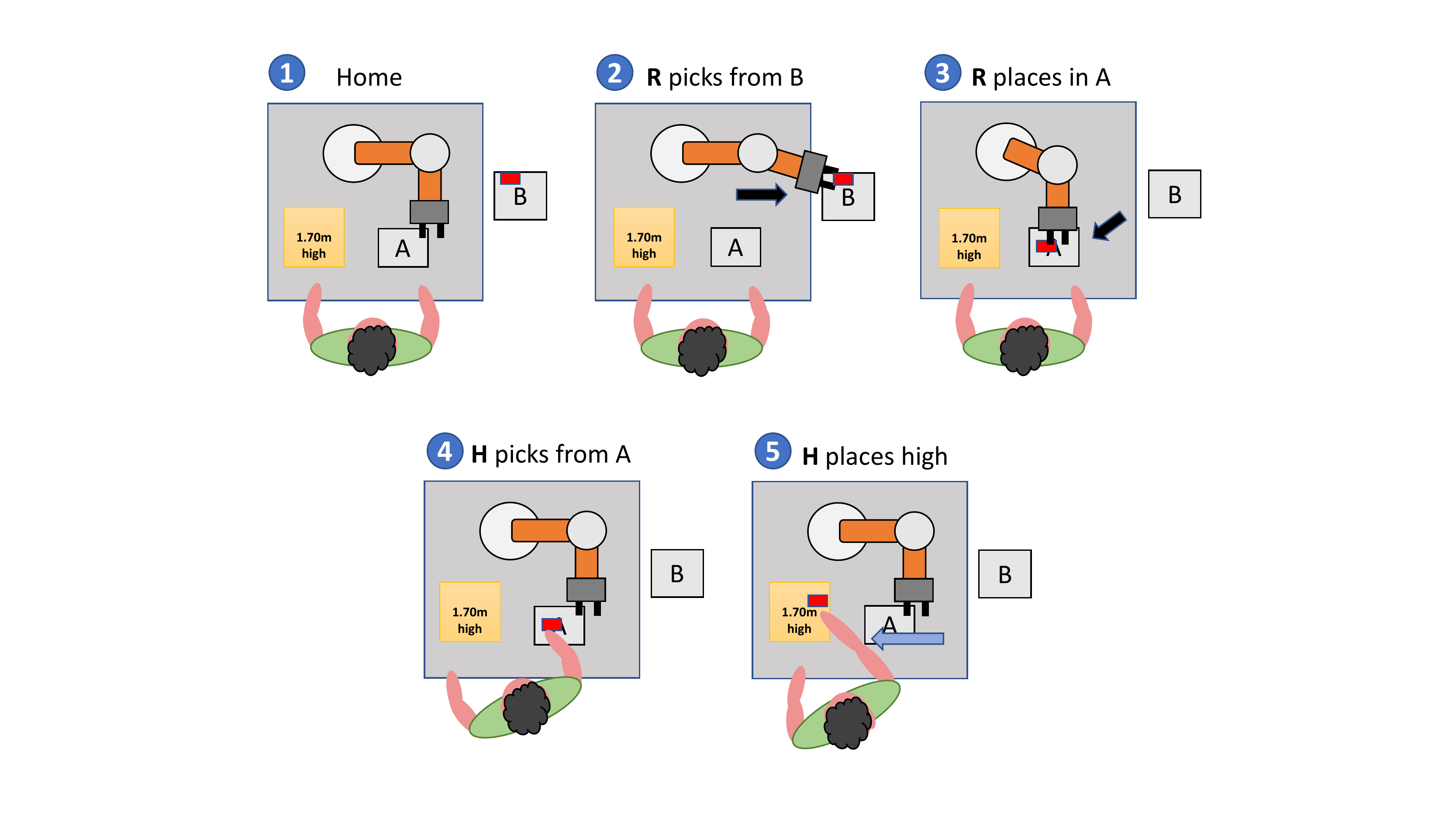}
    \caption{\emph{\textbf{High shelf lifting}} illustration; ``\emph{\textbf{H}}'' stands for ``human operator'', ``\emph{\textbf{R}}'' for ``robot''.}
    \label{fig:span_light}
\end{figure}
\newpage
\begin{itemize}
    \item \emph{\textbf{Hammering}} (\emph{Hammer}). The operator hits with a hammer a metallic tide held by the robot. In this case, the interest was to check how much the collision detection is robust to an action where the human arm is colliding close to the robotic arm (that is, on the metallic tile) without properly colliding \emph{with the robotic arm}.
\end{itemize}
 \begin{figure}[h]
    \centering
  \includegraphics[width=1.0\linewidth]{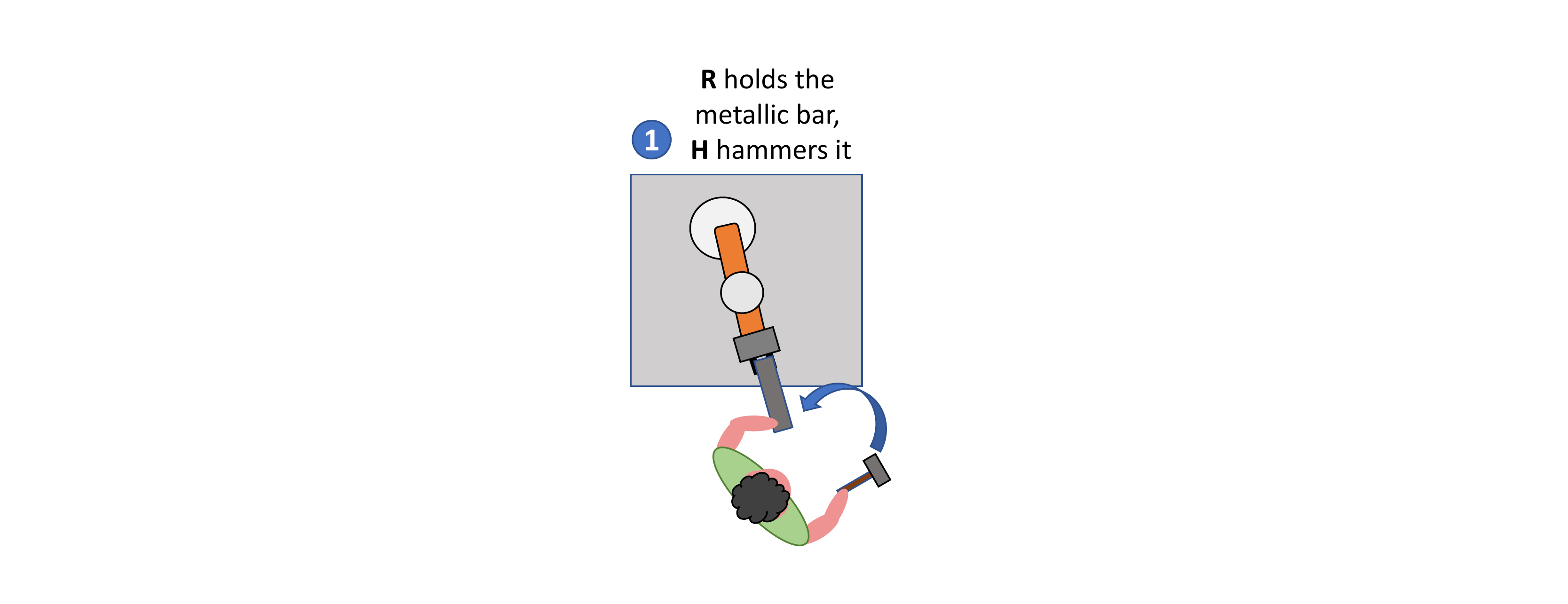}
    \caption{\emph{\textbf{Hammering}} illustration; ``\emph{\textbf{H}}'' stands for ``human operator'', ``\emph{\textbf{R}}'' for ``robot''. This action is important since it tests the collision prediction task against false positive alarms. In facts, this action requires the human to be very close to the robot, keeping the item to hammer with one hand, the other doing the hammering action. In facts, this action was creating the most false positive alarms, around 80\%.}
    \label{fig:span_light}
\end{figure}
\newpage

\subsection{Implementation and Learning}
\label{sec:impl_det}

Further to what included in Secs.~3 and 5 \textit{(main paper)}, we 
discuss here a few additional details on implementation and learning.

\subsubsection{Implementation details.}
The proposed SeS-GCN is written in Pytorch (the source code will be distributed).
The model adopts residual connections at each GCN layers, it is regularized with batch normalization~\cite{batchnorm15} at the end of each GCN layer, and it is optimized with ADAM~\cite{adam15}.

\subsubsection{Learning time.}
On Human3.6M~\cite{h36m_pami}, training of SeS-GCN proceeds for 60 epochs for both the teacher and the student models. We used batch size of 256, learning rate of 0.1, and decay rate of 0.1 at epochs 5, 20, 30 and 37.
On an Nvidia RTX 2060 GPU, the learning process takes 30 minutes.

\subsubsection{Loss function.}

Following literature~\cite{Mao19LTD,Mao20DCT,Dang21}, the loss function differs from the test metric, Eq.~(5) main paper; namely, the loss function considers the average of MPJPEs over the entire predicted sequence:
\begin{equation*}
    L_{\text{MPJPE}}= \frac{1}{VT} \sum_{t=0}^{T} \sum_{v=1}^{V}||\hat{\boldsymbol{x}}_{vt}-\boldsymbol{x}_{vt} ||_{2}
\end{equation*}
where, in accordance with Eq.~(5), $\hat{\boldsymbol{x}}_{vt}$ and $\boldsymbol{x}_{vt}$ are the 3-dimensional vectors of a target joint $j_v\ (0\leq v \leq V)$ in a fixed frame $f_t\ (0 \leq t \leq T)$ for the ground truth and the predictions, respectively.

\begin{figure}[t]
    \centering

    \includegraphics[width=1\linewidth, height=0.56\linewidth]{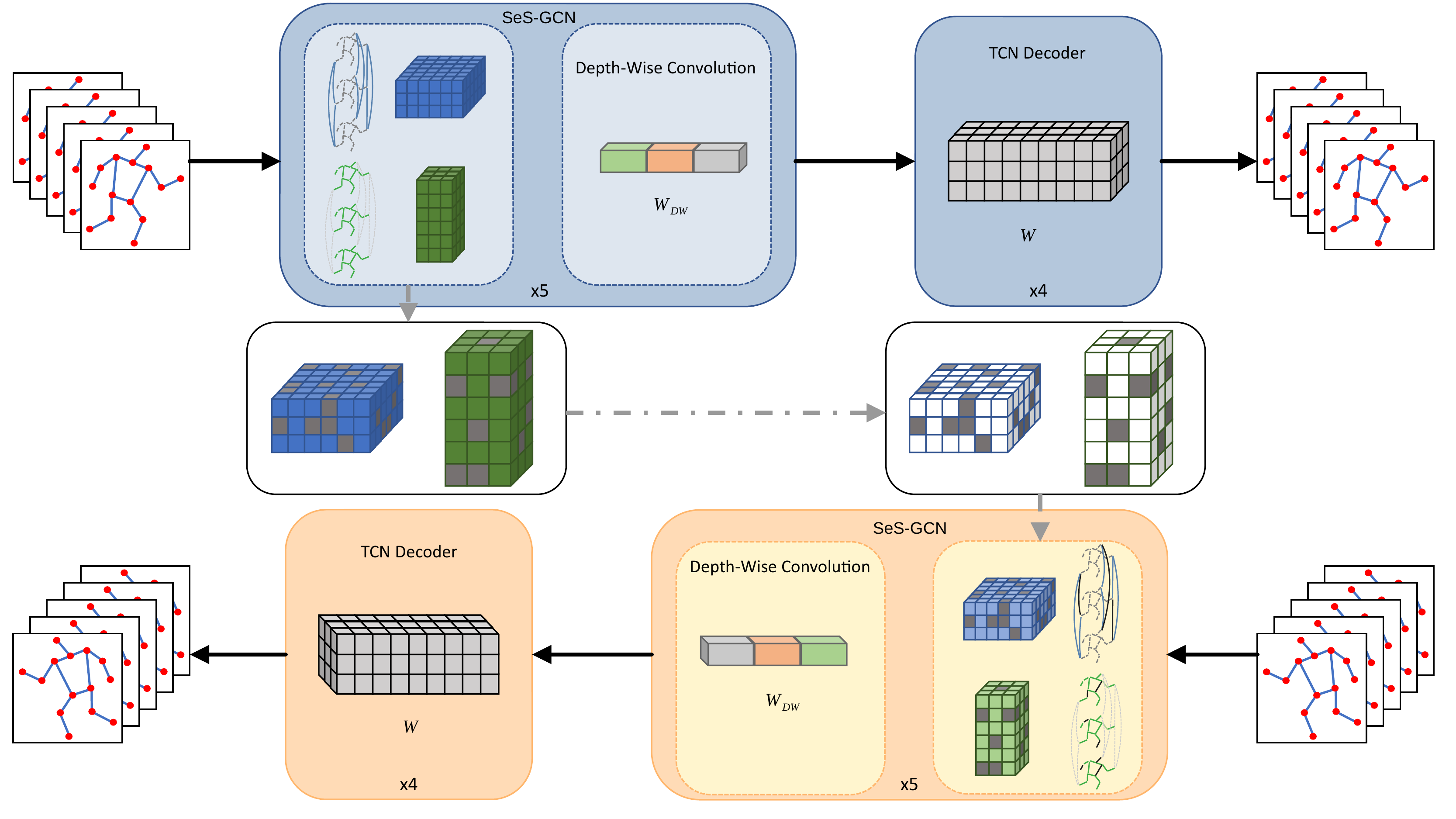}
    \caption{Overview of the proposed pipeline. Given a sequence of observed 3D poses, the Teacher network (depicted with blue boxes) encodes the spatio-temporal body dynamics with 5 SeS-GCN layers, composed by the space-time separable encoder followed by the Depth-Wise convolution. The future trajectories are then predicted with 4 TCN layers. After the train of the Teacher, we threshold the values of the spatial and temporal adjacency matrix to obtain the masks which are then applied during the Student model (depicted in orange boxes) training.}
    \label{fig:error}
\end{figure}

\clearpage
%
%
\bibliographystyle{splncs04}
\bibliography{egbib}
\end{document}